\documentclass[lettersize,journal]{IEEEtran}
\usepackage{graphicx} 
\usepackage{subfigure}
\usepackage{float}
\usepackage{amsmath} 
\usepackage{amssymb}  
\usepackage[noend]{algpseudocode}
\usepackage{algorithmicx,algorithm}
\usepackage{algorithm}
\usepackage{algorithmicx}
\usepackage{algpseudocode}
\usepackage{booktabs}
\usepackage{cite}
\usepackage{threeparttable}
\usepackage{multicol}
\usepackage{multirow}
\usepackage[dvipsnames]{xcolor}
\usepackage{colortbl}
\usepackage{mathtools}
\usepackage{mathrsfs}
\usepackage{amsthm}
\usepackage[colorlinks=true,
            linkcolor=red,
            anchorcolor=blue,
            citecolor=green,
            ]{hyperref}
\usepackage{color}
\usepackage{bm}
\usepackage{tikz}
\usepackage{balance}
\usepackage{colortbl}
\newcommand{\best}{\cellcolor{fcolorFst}\bf}
\newcommand{\sbest}{\cellcolor{fcolorSnd}}
\newcommand{\tbest}{\cellcolor{fcolorTrd}}
\colorlet{fcolorFst}{Green!25}
\colorlet{fcolorSnd}{SpringGreen!45}
\colorlet{fcolorTrd}{Yellow!30}

\begin{document}

\title{ Decoupled Rotation and Translation Estimation from Triple Point-line Images for Visual Odometry}

\author{Zewen Xu$^{1,2}$, Yijia He$^{3}$, Hao Wei$^{1*}$, Bo Xu$^{4}$, BinJian Xie$^{1,2}$ and Yihong Wu$^{1,2*}$
\thanks{*This work was supported by Beijing Science and Technology Plan Project Z231100007123005 and a
SINOPEC Research Project. (Corresponding authors: Hao Wei and Yihong Wu.)}
\thanks{$^{1,2}$Zewen Xu, Hao Wei, Binjian Xie, and Yihong Wu are with the State Key Laboratory of Multimodal Artificial Intelligence Systems, Institute of Automation, Chinese Academy of Sciences, Beijing 100190, China. Zewen Xu, Binjian Xie and Yihong Wu are also with the School of Artificial Intelligence, University of Chinese Academy of Sciences, Beijing 100190, China (e-mail: \{xuzewen2020; weihao2019; xiebinjia2022; yihong.wu\}@ia.ac.cn).}
\thanks{$^{3}$ Yijia He is with 
TCL RayNeo, China (e-mail: heyijia2016@gmail.com)}
\thanks{$^{4}$ Bo Xu is with the School of Geodesy and Geomatics,
Wuhan University, Wuhan 430079, China (e-mail: boxu1995@whu.edu.cn).}
}

\markboth{Journal of \LaTeX\ Class Files,~Vol.~14, No.~8, August~2021}%
{Shell \MakeLowercase{\textit{et al.}}: A Sample Article Using IEEEtran.cls for IEEE Journals}


\maketitle

\begin{abstract}
Line features are valid complements for point features in man-made environments. 3D-2D constraints provided by line features have been widely used in Visual Odometry (VO) and Structure-from-Motion (SfM) systems. However, how to accurately solve three-view relative motion only with 2D observations of points and lines in real time has not been fully explored, which is important for the initialization process of VO systems. In this paper, we propose a novel three-view pose solver based on rotation-translation decoupled estimation. First, a high-precision rotation estimation method based on normal vector coplanarity constraints that consider the uncertainty of observations is proposed, which can be solved by Levenberg-Marquardt (LM) algorithm efficiently. Second, a robust linear translation constraint that minimizes the degree of the rotation components and feature observation components in equations is elaborately designed for estimating translations accurately.
Experiments on synthetic data and real-world data show that the proposed approach improves both rotation and translation accuracy compared to the classical trifocal-tensor-based method and the state-of-the-art two-view algorithm in outdoor and indoor environments.
\end{abstract}


\section{Introduction}
\begin{figure}[t]
\centering
\begin{minipage}{0.24\linewidth}
\centering
    \includegraphics[height=0.79\linewidth]{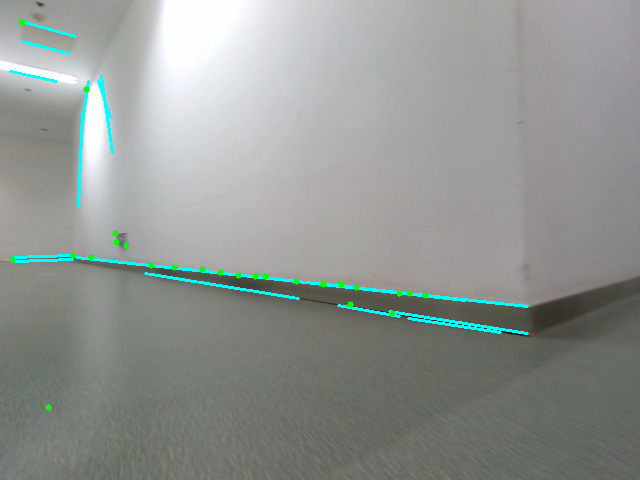}
    \includegraphics[height=0.79\linewidth]{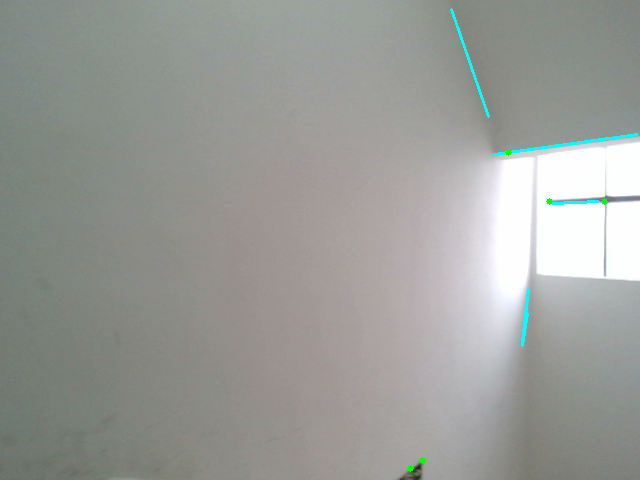}
    \includegraphics[height=0.79\linewidth]{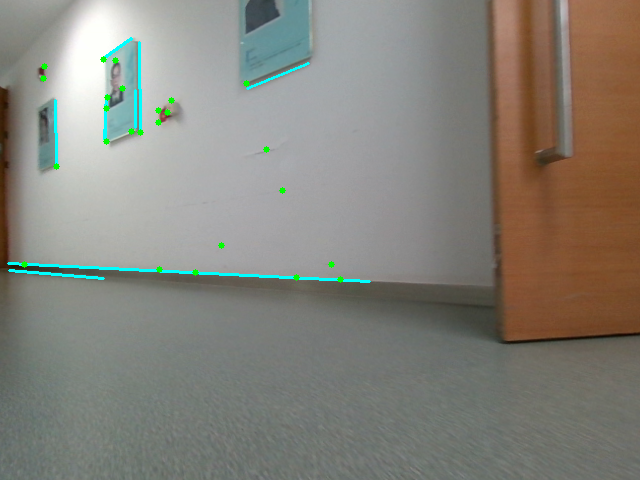}
    \includegraphics[height=0.79\linewidth]{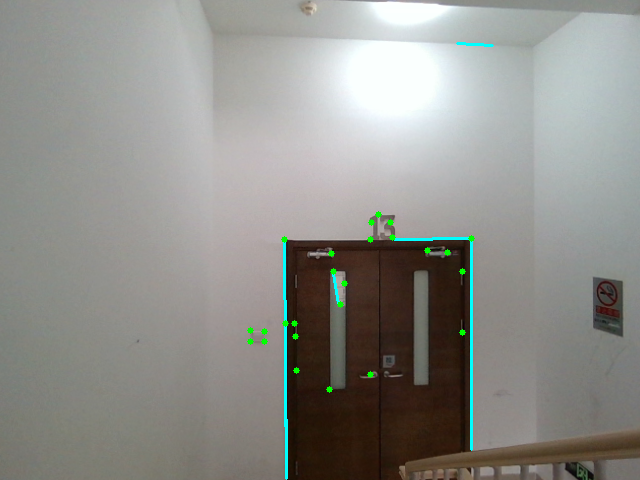}
\end{minipage}
\begin{minipage}{0.24\linewidth}
\centering
    \includegraphics[height=0.79\linewidth]{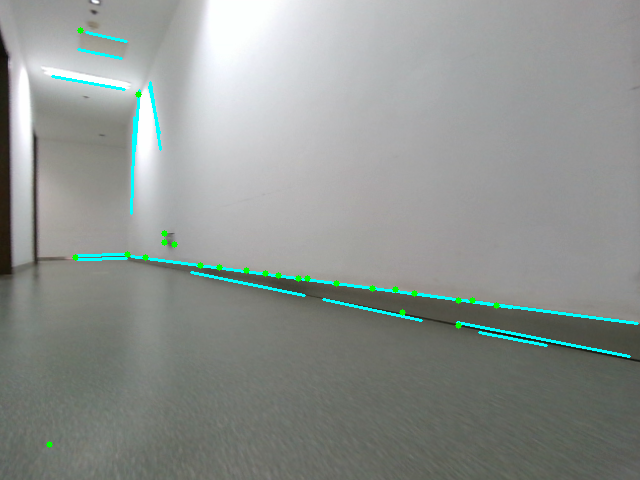}
    \includegraphics[height=0.79\linewidth]{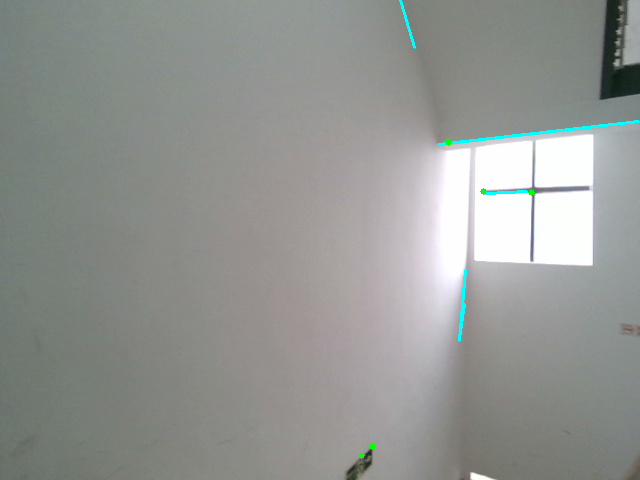}
    \includegraphics[height=0.79\linewidth]{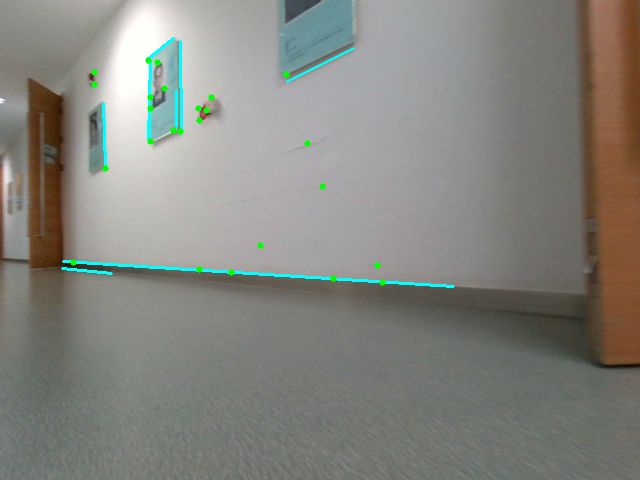}
    \includegraphics[height=0.79\linewidth]{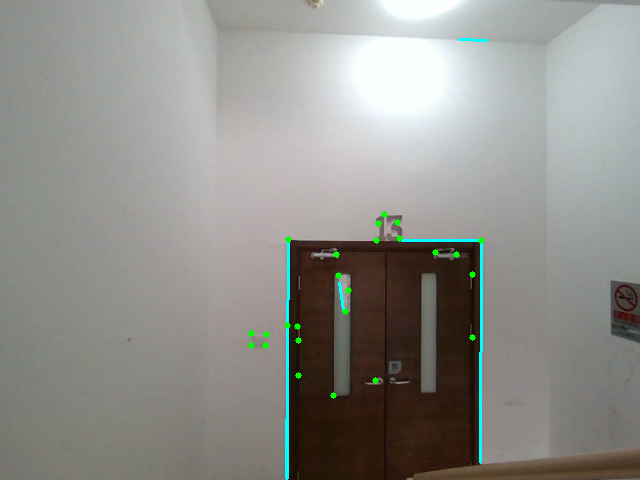}
\end{minipage}
\begin{minipage}{0.24\linewidth}
\centering
    \includegraphics[height=0.79\linewidth]{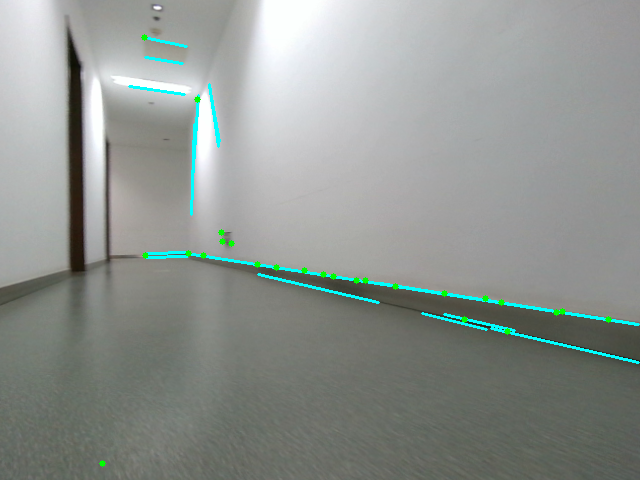}
    \includegraphics[height=0.79\linewidth]{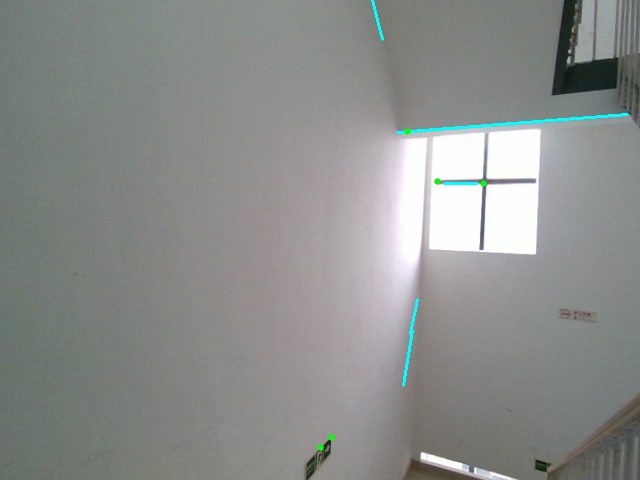}
    \includegraphics[height=0.79\linewidth]{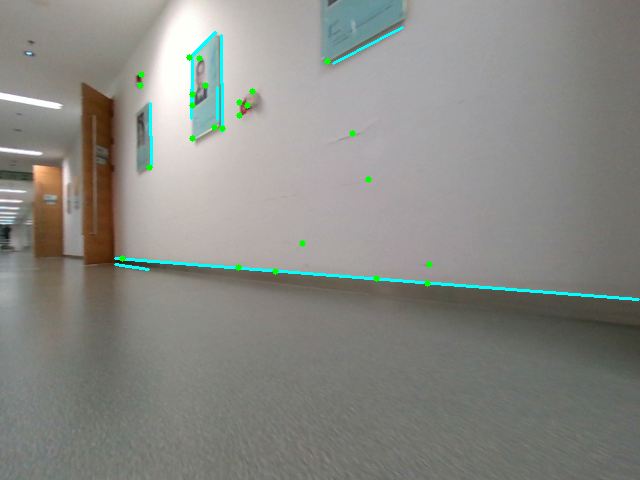}
    \includegraphics[height=0.79\linewidth]{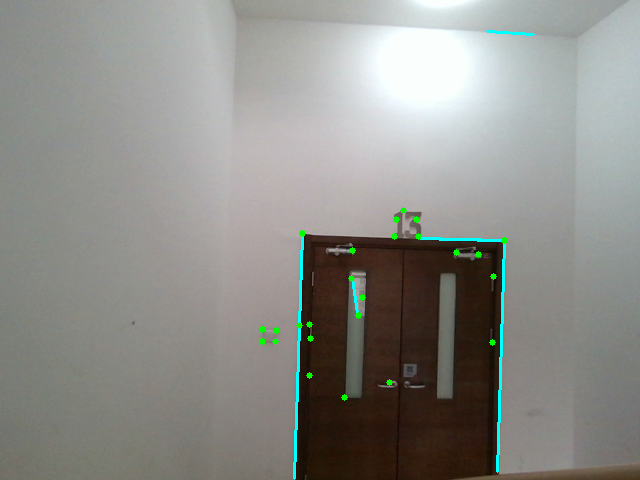}
\end{minipage}
\begin{minipage}{0.24\linewidth}
\centering
\includegraphics[height=0.79\linewidth]{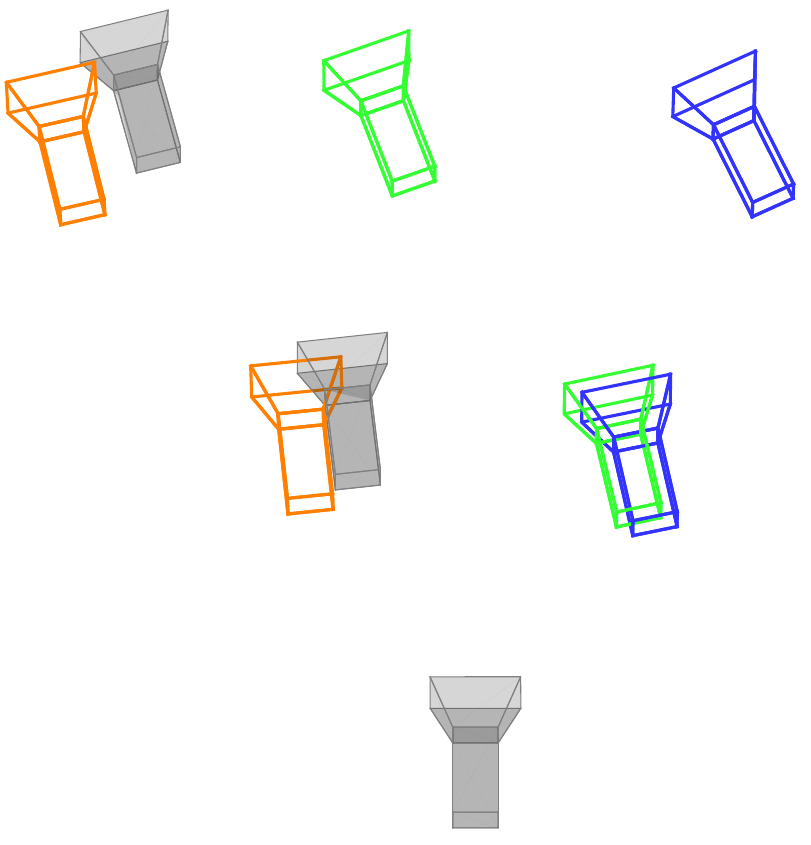}
\includegraphics[height=0.79\linewidth]{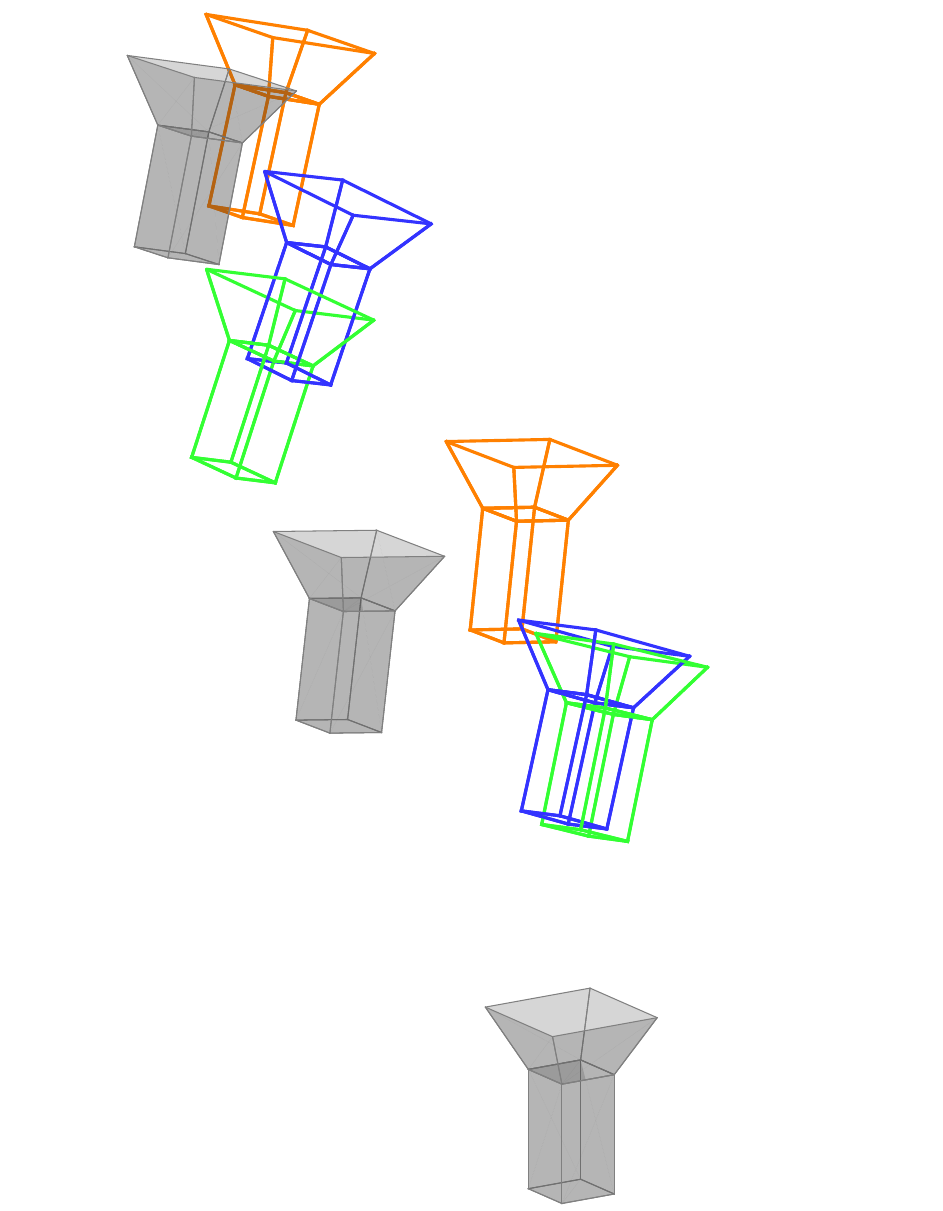}
\includegraphics[height=0.79\linewidth]{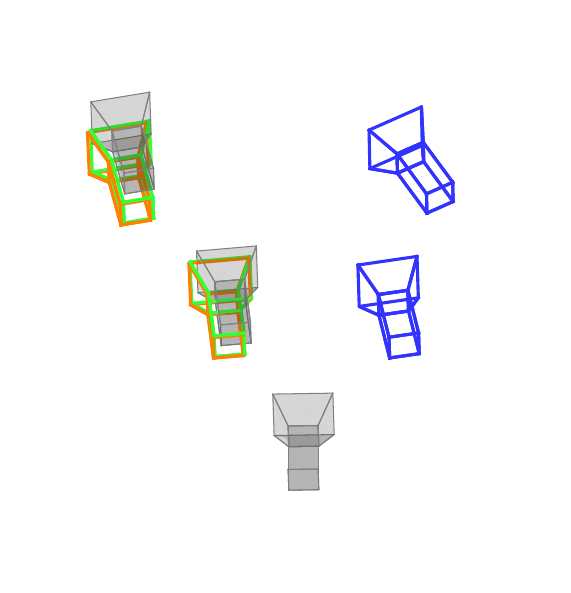}
\includegraphics[height=0.79\linewidth]{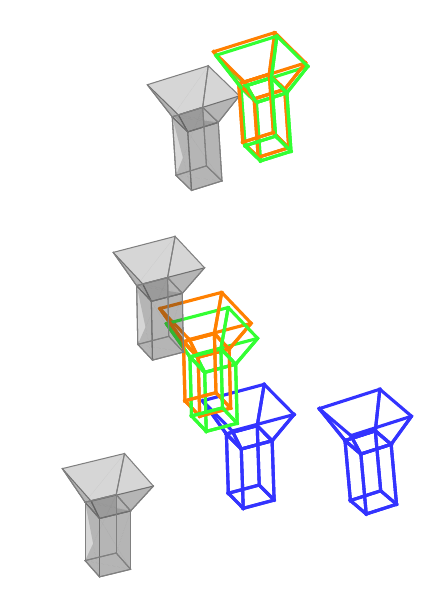}
\end{minipage}

\caption{\textbf{Results of the proposed method test on some difficult scenes.} The first three columns of pictures represent the triple point-line images, where green points denote matched 2D point features and cyan lines denote matched 2D line features. The pose estimation results are shown on the right. Ground truth poses are in solid black, estimated poses of the proposed method (RT$^2$PL) are in orange, estimated poses of PNEC \cite{muhle2022probabilistic} are in green, and estimated poses of five point method \cite{stewenius2006recent} are in blue. All of estimated translation are recovered with true scale for comparisons. The cases shown above are selected from CID-SIMS datasets \cite{zhang2023cid}, where the first row and third row are sampled form Seq. floor3\_1, the second row and the forth row are sampled from Seq. 14-13-12. The first two rows show the performance of pose estimation methods in weak-texture cases. The last two rows show the performance of pose estimation methods in planar degeneracy cases.
}
\label{fig:special_cases}
\end{figure}

\IEEEPARstart{L}{ine} features have been proved as valid complements for point features, particularly in weak-texture scenes \cite{yammine2014novel,li2018reliable}.
They mainly contribute to the robustness of pose estimation with the correspondences between 3D landmarks and 2D observations \cite{liu1990determination, vakhitov2021uncertainty,zhang2016comparative} in most point-line-based VO systems \cite{gomez2019pl, he2018pl, wei2021point, wei2022structural} or SfM systems \cite{bartoli2005structure, hofer2017efficient, wei2022elsr, liu20233d}. However, at the start of these systems, there are no 3D structures before triangulation. It means that an initial pose must be estimated only with 2D correspondences in images. 
Although a non-real-time pose solver with 2D points and lines has been proposed to address weak-texture cases in SfM applications \cite{fabbri2020trplp}, only point features are used to obtain the initial pose in most existing point-line-based VO systems \cite{gomez2019pl, he2018pl, wei2021point, wei2022structural} due to the lack of an accurate and real-time 2d-point-line-based pose estimator. There is no doubt that this absence will limit further applications of line features in VO systems, especially in weak-texture scenes, as shown in the first two rows of Fig. \ref{fig:special_cases}.

In existing VO systems, the essential matrix is widely used in calibrated two-view cases. Through the essential matrix, poses can be estimated accurately and efficiently with at least five points \cite{nister2004efficient,izquierdo2003estimating,stewenius2006recent}. However, these methods suffer from problems caused by the mixing parameters of rotation and translation, including solution multiplicity, planar degeneracy, and pure rotation degeneracy \cite{kneip2012finding, kneip2013direct}. For example, the five-point method cannot offer accurate pose estimation when points distribute on a plane as shown in the last two rows of Fig. \ref{fig:special_cases}. To settle these problems, Kneip et al. \cite{kneip2012finding} proposed a solution that achieved decoupled estimation of rotation and translation by introducing an equivalent constraint to the essential matrix, known as the normal epipolar constraint (NEC). Their later work \cite{kneip2013direct} provided a real-time solver for NEC. Besides, Muhle \textit{et al.} \cite{muhle2022probabilistic} considered the position uncertainty of points and proposed a solution for the NEC problem with iteratively reweighted least square (IRLS) algorithm  \cite{lawson1961contribution}, called PNEC. Recently, due to the advantages offered by decoupled estimation,  NEC-based methods gradually get more attention and replace essential-matrix-based methods in odometry systems \cite{muhle2022probabilistic, chng2020monocular, concha2021instant, he2023rotation, wang2024stereo}. However, there are few attentions on decoupled estimation with point-line images. \textit{To enhance the robustness of NEC-based methods in weak-texture environments, we extend the decoupled estimation to point-line cases.}

\begin{figure*}[t]
\centering
  \subfigure[Normal Epipolar Constraint (NEC)]{
     \includegraphics[width=0.23\linewidth]{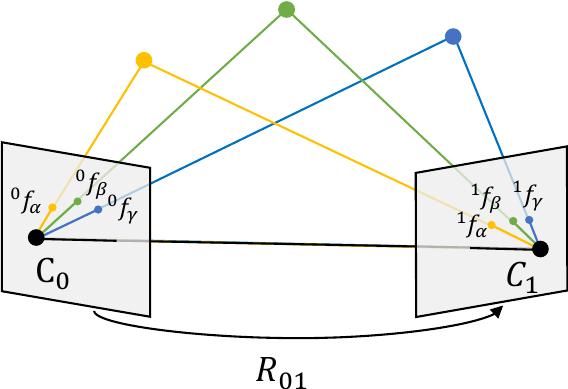}
     \includegraphics[width=0.23\linewidth]{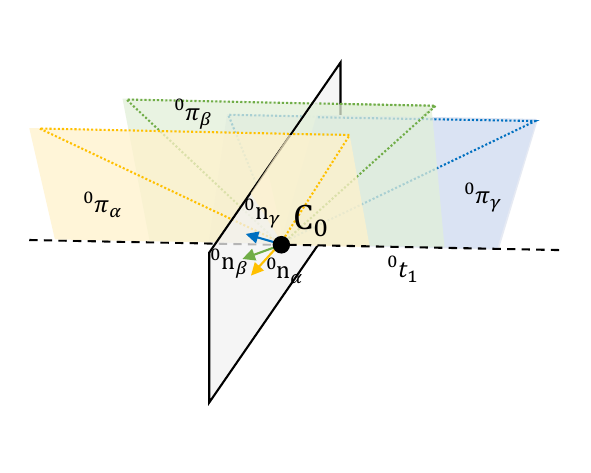}
     \label{fig:point_constraint}
  }
  \subfigure[Normal Back-projected Constraint (NBC)]{
     \includegraphics[width=0.23\linewidth]{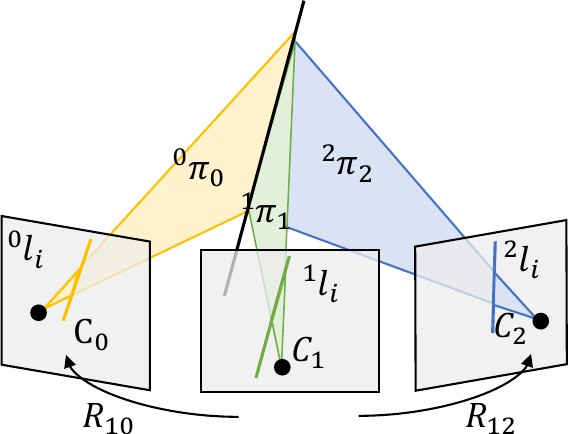}
     \includegraphics[width=0.23\linewidth]{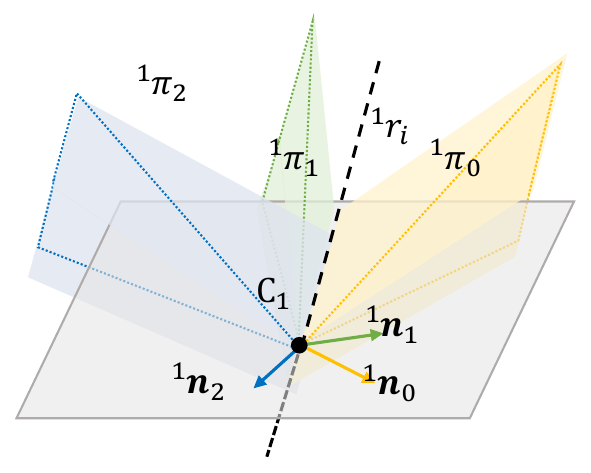}
     \label{fig:line_constraint}
  }
   \caption{\textbf{Geometry of the constraints about rotations.} (a) \textbf{NEC}: For clarity, we only show a constraint provided by three point correspondences ($\alpha$, $\beta$, and $\gamma$) in two frames. In the left picture, the projections of the 3D point $\alpha$ (yellow point) on the two images are represented as ${}^{0}\boldsymbol{f}_\alpha$ and ${}^{1}\boldsymbol{f}_\alpha$, respectively. 
   It is obvious that the plane span by point $\alpha$, camera optical center $\boldsymbol{C}_0$ and camera optical center $\boldsymbol{C}_1$ is passing through the translation vector. This plane, namely \textit{the epipolar plane}, can be represented in the frame 0 as ${}^{0}\boldsymbol{\pi}_\alpha$ , shown as the yellow plane in the right picture, whose normal vector ${}^{0}\boldsymbol{n}_\alpha$ can be obtained by Eq. \ref{eq:noraml_point}. The plane ${}^{0}\boldsymbol{\pi}_\beta$ and the plane ${}^{0}\boldsymbol{\pi}_\gamma$ also pass through the translation vector, which means their normal vectors are coplanar. According to Eq. \ref{eq:noraml_point},  Eq. \ref{eq:NEC_M} and Eq. \ref{eq:point_eigenvalue}, this constraint is only about the relative rotation $\boldsymbol{R}_{01}$.
   (b) \textbf{NBC}: For clarity, we only show a constraint provided by a line across three frames. \textit{The back-projected plane} is defined by the 2D line observation and the camera optical center like ${}^0\pi_0$, ${}^1\pi_1$ and ${}^2\pi_2$ in the left picture. These back-projected planes of the 2D line correspondences are intersecting with the related 3D line. Therefore, the normal vectors of these back-projected planes are coplanar. We represent these three normal vectors in the Frame 1 as ${}^{1}\boldsymbol{n}_0$, ${}^{1}\boldsymbol{n}_1$ and ${}^{1}\boldsymbol{n}_2$ as shown in the right picture, which can be obtained by  Eq. \ref{eq:back-projected_normal}. This constraint is only about the relative rotations $\boldsymbol{R}_{10}$ and $\boldsymbol{R}_{12}$. Therefore, whatever for points and lines, the rotations can be estimated decoupled with translations through NEC and NBC.
   }
\end{figure*}

Pose estimation with point-line images is not a new topic.
When taking line features into account without any priors, it is necessary to consider observations from at least three images. This is because the degrees of freedom of a 3D line are equal to the constraints provided by the observations in two-view cases, resulting in no possibility of removing the effects of measurement errors \cite{hartley2003multiple}. Three-view pose estimation is believed as a fallback when two-view pose estimation fails \cite{guan2022trifocal}. 
In theory, 6 lines or 4 points are enough to solve the calibrated three-view pose problems, known as trifocal minimal solutions \cite{hartley2003multiple}. However, these solvers yield a large number of spurious solutions \cite{holt1994motion}, thus believed as hard minimal problems \cite{hruby2022learning}. The complete classification of trifocal minimal problems can be found in \cite{duff2019plmp,fabbri2022trifocal}.
In addition to facing spurious solutions, different point-line configurations also need to be discussed separately, further increasing the difficulties in practical applications. Fabbri \textit{et al.} \cite{fabbri2020trplp} addressed two trifocal minimal problems, utilizing the homotopy continuation algorithm \cite{morgan2009solving} to settle the spurious-solution problem and enhance efficiency. However, this method still spends hundreds of milliseconds on a slove process  (660ms on average on an Intel core i7-7920HQ processor with four threads reported in \cite{fabbri2020trplp}), which limits its application in odometry systems. 

When considering more features, the solving process becomes much easier. A classical approach is the trifocal-tensor-based method \cite{hartley2003multiple}, which can lead to a real-time linear solver with no fewer than 13 lines, 7 points, or a mixture of both. Since all features are considered together, when the observations are sufficient, the solver can avoid the complex process of distinguishing the exact point-line configuration. This characteristic is obvious but important for the efficiency of its practical application in odometry systems.
However, the trifocal-tensor-based method overlooks the non-linear internal constraints of rotations, consequently failing to deliver satisfactory accuracy. Additionally, this method cannot handle planar or pure rotation degeneracy due to the mixing parameter problem, as will be shown in Sec \ref{sec:Noise_Resilience}. Therefore, this method is also not used in most point-line-based VO systems.

To this end, our primary motivation is to design an accurate and real-time pose estimation with point-line images for odometry applications. This algorithm should be capable of handling situations involving pure rotation or planar degeneracy. The contributions of this paper are summarized as follows:
\begin{itemize}
\item An accurate and \textbf{R}eal-\textbf{T}ime pose estimator with \textbf{P}oint-\textbf{L}ine trifocal images is proposed, which estimates \textbf{R}otation and \textbf{T}ranslation separately, allowing it to handle pure rotation degeneracy and planar degeneracy, namely \textbf{RT$^2$PL}.
\item Two forms of coplanarity constraints for rotation estimation with lines are proposed. We prove that one form is superior to another on convergence through experiments (see Sec. \ref{sec:convergence}). Then the better one is combined with NEC leading to a point-line-based rotation solver, which can be easily solved by LM algorithm in real time. Additionally, the cost function integrates position uncertainty of points and lines, improving resilience to noise.
\item A novel point-line-based linear translation constraint is introduced, meticulously designed to minimize the degree of the product of rotation and the degree of the product of feature observations, thereby enhancing the resilience against rotation estimation errors and observation noise (see Sec. \ref{sec:ligt_synthic}).
\item  Extensive experiments on synthetic and real-world datasets demonstrate the effectiveness of RT$^2$PL, which outperforms both the classical trifocal-tensor-based method and the state-of-the-art point counterparts in general and degeneracy cases. 
\end{itemize}

The paper is organized as follows. Sec. \ref{sec:rotation} introduces the rotation estimation part of RT$^2$PL. Sec. \ref{sec:translation} introduces the translation estimation part of RT$^2$PL. The performance of proposed methods is demonstrated in Sec. \ref{sec:experiments}. The paper is concluded in Sec. \ref{sec:conclusion}. 

\section{Rotation Estimation}
\label{sec:rotation}
NEC \cite{kneip2013direct} constructed by point observations leads to an estimator where the rotation is recovered in a decoupled way. Inspired by it, for line features, normal back-projected constraint (NBC) is proposed in this paper. Integrating them and considering observation uncertainty results in a probability-aware point-line-based rotation estimation method, which is the rotation estimation part of RT$^2$PL. 
\subsection{Background-NEC}
\label{sec:background}
The observations in two images of a 3D points are set as ${}^k\boldsymbol{f}$, $k = 0,1$, whose homogeneous coordinates ${}^k\overline{\boldsymbol{f}}$ are
\begin{equation}
    {}^k\overline{\boldsymbol{f}} = \begin{bmatrix}
        {}^ku&{}^kv&1
    \end{bmatrix}^T,\quad k = 0,1.
\end{equation} 
The unit bearing vector corresponding to ${}^k\boldsymbol{f}$ is
\begin{equation}
    {}^k\bold{b} = \frac{\bold{K}^{-1}{}^k\overline{\boldsymbol{f}}}{\left\|\bold{K}^{-1}{}^k\overline{\boldsymbol{f}}\right\|_2},
    \label{eq:bearing_vector}
\end{equation}
where $\bold{K}$ is the camera intrinsics matrix. The epipolar plane corresponding to the 3D point is defined by its two bearing vectors. The normal vector of the epipolar plane represented in frame $0$ is obtained by 
\begin{equation}
    {}^0\bold{n} = {}^0\bold{b} \times \bold{R}_{01}{}^1\bold{b}
    \label{eq:noraml_point}
\end{equation}
where $\bold{R}_{ab} \in SO(3)$ denotes the rotation that takes 3D features from frame $b$ to frame $a$. 
The normal vectors of all these epipolar planes span a plane orthogonal to the translation vector between two frames. The unit translation vector represented in frame 0 is ${}^0\bold{t}_1$ (see the right picture of Fig. \ref{fig:point_constraint}). NEC constraint enforces the coplanarity of normal vectors of epipolar planes and builds the following error:
\begin{equation}
    {}^0\mathit{e}= \left|{}^0\bold{t}_1^T{}^0\bold{n}\right|.
\end{equation}
Assuming the number of 3D points that are observed by the two frames is $m$, the energy function is
\begin{equation}
    E_{point} = \sum\limits_{j=1}^{m}{}^0e_{j}^2 = {}^0\bold{t}_1^T{}^0\bold{M}{}^0\bold{t}_1,
    \label{eq:point_cost_function}
\end{equation}
where
\begin{equation}
{}^0\bold{M}=\sum\limits_{j=1}^{m}{}^0\bold{n}_j{}^0\bold{n}_j^T.
\label{eq:NEC_M}
\end{equation}
Minimizing Eq. (\ref{eq:point_cost_function}) is equal to minimizing the minimal eigenvalue of ${}^0\bold{M}$, denoted by $\lambda_{{}^0\bold{M}}^{min}$,
\begin{equation}
\bold{R}_{01}=
    \underset{\bold{R}_{01}}{\arg\min}E_{point}=  \underset{\bold{R}_{01}}{\arg\min}\lambda_{{}^0\bold{M}}^{min}.
    \label{eq:point_eigenvalue}
\end{equation}
For more detailed derivation and geometric interpretations, we refer readers to the original paper \cite{kneip2013direct} and Fig. \ref{fig:point_constraint}. Because the minimal eigenvalue is only related to rotation parameters, the rotation can be estimated in a decoupled way. 
\subsection{Normal Back-projected Constraint}
\label{sec:NBC}
We propose NBC for rotation estimation with lines. The geometry of this constraint is shown in Fig. \ref{fig:line_constraint}. 2D observation of a 3D line in the $k$-th image  is represented as ${}^k\boldsymbol{l}$ with its polar coordinates $[{}^k{\rho},{}^k{\theta}]$,
\begin{equation}
  {}^k\boldsymbol{l} = \begin{bmatrix}\sin{{}^k{\theta}} & -cos{{}^k{\theta}} & {}^k{\rho} \end{bmatrix}^T.
  \label{eq:line_rep}
\end{equation}
The back-projected plane that passes through the $k$-th camera optical center
and ${}^k\boldsymbol{l}$ is represented as  ${}^k\boldsymbol{\pi}_{k}= [{}^k\bold{n}_{k}^T, 0]^T$, where ${}^k\bold{n}_{k}$ is normal unit vector of the plane,
\begin{equation}
    {}^k\bold{n}_{k}=\frac{\bold{K}^T{}^k\boldsymbol{l}}{\left \| \bold{K}^T{}^k\boldsymbol{l} \right \|_2}.
    \label{eq:bp_normal}
\end{equation}
If all normal vectors of back-projected planes corresponding to the line landmark are represented in a unified coordinate system (frame $1$ for example), 
\begin{equation}
{}^1\bold{n}_k = \bold{R}_{1k}{}^k\bold{n}_k,
\label{eq:back-projected_normal}
\end{equation}
they will span a plane orthogonal to the direction of the landmark ${}^1\bold{r}$ (see the right picture of Fig. \ref{fig:line_constraint}). Therefore, the error of the model can be built according to the coplanarity as below,
\begin{equation}
    {}^{1}\mathit{e}_{k} = \left|{}^1\bold{r}^T {}^1\bold{n}_k\right|,
\end{equation}
where ${}^1\bold{n}_k$ denotes the normal vector of the $k$-th back-projected plane represented in frame $1$. Each line correspondence across three frames can provide a coplanarity constraint. Assuming the number of 3D lines that can be observed by three frames is $n$. The energy function can be constructed similarly to NEC as below, 
\begin{equation}
E_{line} =  \sum\limits_{i=1}^{n}\sum\limits_{k=0}^{2} {}^{1}\mathit{e}_{k,i}^2 = \sum\limits_{i=1}^{n}{}^1\bold{r}_i^T {}^1\bold{M}_{i} {}^1\bold{r}_i =\sum\limits_{i=1}^{n}\lambda_{{}^1\bold{M}_{i}}^{min},
\label{eq:minimal_function}
\end{equation}
where
\begin{equation}
         {}^1\bold{M}_{i} = \sum\limits_{k=0}^{2}{}^1\bold{n}_{k,i}{}^1\bold{n}_{k,i}^T.
         \label{eq:M_contraints}
\end{equation}
 We still use $\lambda_{{}^1\bold{M}_{i}}^{min}$ denoting the minimal eigenvalue of ${}^1\bold{M}_{i}$. Enforcing each error item led by $n$ coplanarity constraints to zero, the problem can be solved  by
\begin{equation}
    \left[ \bold{R}_{10} \ \bold{R}_{12} \right] =\underset{ \left[\bold{R}_{10} \ \bold{R}_{12} \right]}{\arg\min}\begin{bmatrix}
        \lambda_{{}^1\bold{M}_{1}}^{min}\\
       \lambda_{{}^1\bold{M}_{2}}^{min}\\
        \vdots\\
        \lambda_{{}^1\bold{M}_{n}}^{min}
    \end{bmatrix}.
    \label{eq:minimal_form}
\end{equation}
It is worth noticing that, under the aforementioned constraints, there is no need to have any prior information about 3D lines, such as the directions of lines, which are the eigenvectors related to those minimal eigenvalues. 

\textbf{Another Form for NBC: } The coplanarity constraint can be built in another error form:
\begin{equation}
\begin{aligned}
        {}^{1}\mathit{e}_{i} &= \left|{}^1\bold{n}_0^T( {}^1\bold{n}_{1} \times  {}^1\bold{n}_2)\right| \\
        &= \left|\left(\begin{vmatrix} 
         {}^1\bold{n}_0 & {}^1\bold{n}_1 & {}^1\bold{n}_2
        \end{vmatrix} \right)\right|\\
        &= \left|\left(\begin{vmatrix}{}^1\bold{N}_i\end{vmatrix} \right)\right|,
\end{aligned}
\end{equation}
where ${}^1\bold{N}_i$ is a matrix constructed by ${}^1\bold{n}_0$, ${}^1\bold{n}_1$, and ${}^1\bold{n}_2$. It has the following relationship with ${}^1\bold{M}_i$ in Eq. (\ref{eq:M_contraints}):
\begin{equation}
    {}^1\bold{M}_i = {}^1\bold{N}_i^T{}^1\bold{N}_i.
\end{equation}
Therefore, 
\vspace{-0.05in}
\begin{equation}
        {}^{1}\mathit{e}_{i} = \sqrt{\begin{vmatrix}
            {}^1\bold{M}_i
        \end{vmatrix}}= \sqrt{ \prod\lambda_{{}^1\bold{M}_{i}}}.
\end{equation}
The sum of squares of these residuals is given by
\begin{equation}
E_{line}=\sum\limits_{i=1}^{n}{}^{1}\mathit{e}_i^2=\sum\limits_{i=1}^{n}(\prod\lambda_{{}^1\bold{M}_{i}}).
\label{eq:multiple_function}
\end{equation}
Enforcing each error item to zero, the problem is built as
\begin{equation}
    \left[ \bold{R}_{10} \ \bold{R}_{12} \right] =\underset{ \left[\bold{R}_{10} \ \bold{R}_{12} \right]}{\arg\min}\begin{bmatrix}
        \prod\lambda_{{}^1\bold{M}_{1}}\\
        \prod\lambda_{{}^1\bold{M}_{2}}\\
        \vdots\\
        \prod\lambda_{{}^1\bold{M}_{n}}
    \end{bmatrix}.
    \label{eq:multiply_form}
\end{equation}

We call these two forms of NBC as minimal eigenvalue form (Eq. (\ref{eq:minimal_form}), \textbf{NBC-mini)} and eigenvalue multiplication form (Eq. (\ref{eq:multiply_form}), \textbf{NBC-mult}), respectively. The geometric meaning of one item in NBC-mini is the sum of squares of distances between the unit normal vectors and the fitted plane. The geometric meaning of one item in NBC-mult is the volume of the parallelepiped constructed by the three unit normal vectors. The quantitative and qualitative experiments in Sec. \ref{sec:convergence} will show that the NBC-mult form outperforms the NBC-mini form on the behavior of the cost function and the initial value resilience. 
\subsection{Rotation Part of RT$^2$PL}
Considering three frames with $m$ point correspondences and $n$ line correspondences, the total cost function is the combination of Eq. (\ref{eq:point_cost_function}) and Eq. (\ref{eq:multiple_function}):
\begin{equation}
\begin{aligned}
     E(\bold{R}_{10}, \bold{R}_{12}) &= E_{point}(\bold{R}^T_{10}) + E_{point}(\bold{R}_{12})\\
     &+ E_{point}(\bold{R}^T_{10}\bold{R}_{12}) + E_{line}(\bold{R}_{10}, \bold{R}_{12}).
    \label{eq:point_line_cost_function}
\end{aligned}
\end{equation}
Although Eq. (\ref{eq:point_line_cost_function}) enforces the coplanarity of normals of epipolar planes and the coplanarity of normals of back-projected planes, the importance of different features in the cost function is ignored, which has a significant impact on the performance of estimator \cite{muhle2022probabilistic}. 
According to Eq. (\ref{eq:multiple_function}), it is observed that the cost function constructed by lines is in the form of the sum of squared errors, similar to the point-based method Eq. (\ref{eq:point_cost_function}). Each error term is only related to a single correspondence. Therefore, an IRLS version for solving the NBC problem, named PNBC, can be constructed naturally in a similar way to the PNEC \cite{muhle2022probabilistic}. 
To avoid the affection of occlusion and fragmentation problems, we model the uncertainty of line features as our pervious work \cite{xu2023plpl} and set the vertical uncertainty as $\sigma_{line}^2$, then the covariance of line observation is given by
\begin{equation}
    \Lambda_{{}^k\boldsymbol{l}_i} = \begin{bmatrix}
        \frac{2}{c^2}\sigma_{line}^2 & -\frac{2d}{c^2}\sigma_{line}^2 \\
         -\frac{2d}{c^2}\sigma_{line}^2  & \left(\frac{1}{2} + \frac{2d^2}{c^2}\right)\sigma_{line}^2
    \end{bmatrix},
\end{equation}
where $c$ is the line segment length, and $d$ is the distance from the foot point to the midpoint of the line segment. The covariance of the normal vector of the back-projected plane $\Lambda_{{}^k\bold{n}_{k,i}}$ can be given by the unscented transform \cite{uhlmann1995dynamic} according to Eq. (\ref{eq:line_rep}). The weight of the error item related to the $i$-th line is
\begin{equation}
    w_{line,i} =\left(\sigma_{i}^2\right)^{-1}= \left(\sum\limits_{k=0}^2{\frac{\partial e_i}{\partial {}^k\bold{n}_{k,i}}}^T\Lambda_{{}^k\bold{n}_{k,i}}\frac{\partial e_i}{\partial {}^k\bold{n}_{k,i}}\right)^{-1}.
\end{equation}
Considering the point weights $w_{point}$, obtained by the inverse of the variance of residual (Eq. (4) in \cite{muhle2022probabilistic}) together, the IRLS version of Eq. (\ref{eq:point_line_cost_function}) is
\begin{equation}
\begin{aligned}
         E(\bold{R}_{10},\bold{R}_{12})& = \sum\limits_{j=1}^{m_{01}}w^{(01)}_{point,j}{}^1e_{j}^2(\bold{R}^T_{10})\\
         &+\sum\limits_{j=1}^{m_{12}}w^{(12)}_{point,j}{}^1e_{j}^2(\bold{R}_{12}) \\
         & + \sum\limits_{j=1}^{m_{02}}w^{(02)}_{point,j}{}^1e_{j}^2(\bold{R}^T_{10}\bold{R}_{12})\\
         & + \sum\limits_{i=1}^{n} w_{line,i}{}^1e_i^2(\bold{R}_{10},\bold{R}_{12}).
 \end{aligned}
 \end{equation}
All of the weights are initialized as $1$, and when the optimization solved by LM algorithm is convergent, the weight of each item will be updated with new estimated rotations. 
\section{Translation Estimation}
\label{sec:translation}
For both point features and line features, the constraints on translation will be linear when the rotation is known. To build a linear constraint for translation using both points and lines, at least three frames should be considered.
We set the general form of the constraint of translation among three frames as
\begin{equation}
    \bold{B}{}^G\bold{t}_0 + \bold{C}{}^G\bold{t}_1 + \bold{D}{}^G\bold{t}_2 = \bold{0},
    \label{eq:ligt}
\end{equation}
where $\bold{B}$, $\bold{C}$, and $\bold{D}$ denote coefficient matrices defined by rotations and feature correspondences, $^G\bold{t}_0$, $^G\bold{t}_1$, and $^G\bold{t}_2$ denote the global translation about frame $0$, $1$, and $2$ respectively. We extend the concept in \cite{cai2021pose} and call the constraints similar to Eq. (\ref{eq:ligt}) as linear global translation (LiGT) constraints whatever the type of observations. Intuitively, when the degree of rotation components and observation components are lower, the impact of rotation estimation errors and observation noise on translation estimation will be smaller. Therefore, finding a lower degree coefficient matrix for Eq. (\ref{eq:ligt}) in terms of rotation and observation may be beneficial for the stability of translation estimation. It is worth noting that the degree of rotation components here does not represent the number of occurrences of a specific rotation parameter in the coefficient matrices, but rather the total number of occurrences of all rotation parameters in the coefficient matrices. This is just a rough selection strategy, and its rationality will be discussed in Sec. \ref{sec:ligt_synthic}.

\subsection{Point-based LiGT}
\label{sec:p_ligt}
For point correspondences, the relative translation ${}^0\bold{t}_1$ can be given by essential matrix constraints:
\begin{equation}
    {}^0\overline{\boldsymbol{f}}^TE_{01}{}^1\overline{\boldsymbol{f}}={}^0\overline{\boldsymbol{f}}^T[{}^0\bold{t}_1]_{\times}\bold{R}_{01}{}^1\overline{\boldsymbol{f}}={}^0\overline{\boldsymbol{f}}^T[\bold{R}_{01}{}^1\overline{\boldsymbol{f}}]_{\times}{}^0\bold{t}_1=0,
    \label{eq:E_constraints}
\end{equation}
where $E_{01}$ denotes the essential matrix between frame $0$ and frame $1$. The $[\bold{v}]_{\times}$ denotes the skew-symmetric matrix related to the vector $\bold{v}$. 
Different from \cite{cai2021pose}, we just consider the essential matrix constraint of relative translation ${}^0\bold{t}_1$, ${}^0\bold{t}_2$, and ${}^1\bold{t}_2$ together and represent these relative translations with global translations, i.e., ${}^0\bold{t}_1 = \bold{R}_{0 G}({}^G\bold{t}_1 - {}^G\bold{t}_0)$, ${}^0\bold{t}_2 = \bold{R}_{0 G}({}^G\bold{t}_2 - {}^G\bold{t}_0)$, and ${}^1\bold{t}_2 = \bold{R}_{1 G}({}^G\bold{t}_2 - {}^G\bold{t}_1)$.
The coefficient matrices can be obtained by transforming the three essential matrix constraints into the form of LiGT Eq. (\ref{eq:ligt}):
\begin{equation}
    \begin{aligned}
        \bold{B} &= \begin{bmatrix}
            -({}^0\overline{\boldsymbol{f}}^T[\bold{R}_{01}{}^1\overline{\boldsymbol{f}}]_{\times}\bold{R}_{0 G})\\ -({}^0\overline{\boldsymbol{f}}^T[\bold{R}_{02}{}^2\overline{\boldsymbol{f}}]_{\times}\bold{R}_{0 G}) \\ \bold{0}_{1\times3}
        \end{bmatrix}\\
        \bold{C} &= \begin{bmatrix}
            {}^0\overline{\boldsymbol{f}}^T[\bold{R}_{01}{}^1\overline{\boldsymbol{f}}]_{\times}\bold{R}_{0 G}\\ \bold{0}_{1\times3} \\ -{}^1\overline{\boldsymbol{f}}^T[\bold{R}_{12}{}^2\overline{\boldsymbol{f}}]_{\times}\bold{R}_{1 G}
        \end{bmatrix}\\
        \bold{D} &= -(\bold{B} + \bold{C}).   
    \end{aligned}
    \label{eq:newligt_bcd}
\end{equation}
According to Eq. (\ref{eq:newligt_bcd}), the degree of rotation components and observation components in the coefficient are both equal to $2$, which is lower than those in \cite{cai2021pose} with $3$ degrees and $5$ degrees, respectively. The lower degree of feature components and rotation components lead to the more robust LiGT constraint, which will be proved in synthetic scenes in Sec. \ref{sec:ligt_synthic} and in realistic scenes in Sec. \ref{sec:ligt_real}.
\begin{table*}[]
\renewcommand\arraystretch{1.25}
    \centering
    \caption{Relative pose accuracy at different noise levels} 
    \vspace{-0.1in}
    \begin{threeparttable}
     \resizebox{0.98\linewidth}{!}{
    \begin{tabular}{llcccccccc|cccccccc|cccc}
     \toprule[1.0pt]
         &Cases & \multicolumn{8}{c}{General case} & \multicolumn{8}{c}{Planar degeneracy}  & \multicolumn{4}{c}{Pure rotation degeneracy}\\
         &Noise [pix]& \multicolumn{2}{c}{0.5} & \multicolumn{2}{c}{1.0} & \multicolumn{2}{c}{1.5} & \multicolumn{2}{c}{2.0}& \multicolumn{2}{c}{0.0} &\multicolumn{2}{c}{0.5} & \multicolumn{2}{c}{1.0} & \multicolumn{2}{c}{1.5}  & 0.0 & 0.5 & 1.0 & 1.5\\ 
         &Metric [deg]& $e_{rot}$ & $e_{t}$ & $e_{rot}$ & $e_{t}$ & $e_{rot}$ & $e_{t}$ & $e_{rot}$ & $e_{t}$ & $e_{rot}$ & $e_{t}$ & $e_{rot}$ & $e_{t}$ & $e_{rot}$ & $e_{t}$ & $e_{rot}$ & $e_{t}$ & $e_{rot}$ & $e_{rot}$ & $e_{rot}$ & $e_{rot}$\\
         \midrule
        \multirow{6}{*}{\rotatebox[origin=c]{90}{point}} &7pt \cite{hartley2003multiple}/8pt \cite{hartley1997defense} \tnote{$\dagger$}& 0.47 & 1.10 & 0.89 & 2.12 & 1.36 & 3.17 & 1.85 & 4.37 & 23.94 & 110.1 & 25.43 & 108.2 & 25.51 & 108.0 & 25.29 & 110.9 & 305.8 & 0.07 & 0.14 & 0.22 \\
         &5pt-nist \cite{nister2004efficient}  & 3.32 & 4.27 & 3.56 & 5.87 & 4.76 & 6.85 & 4.86 & 7.34 & \sbest6.25 & \sbest14.52 & 4.83 & 12.64 & 6.00 & 14.03 & 5.56 & 12.89 & 1.12 & 0.22 & 0.46 & 0.93\\
         &5pt-stew \cite{stewenius2006recent} & 0.39 & 0.63 & 0.68 & 1.11 & 1.00 & 1.61 & 1.41 & 2.24 & \tbest14.12 & \tbest38.93 & 4.19 & 9.55 & 4.74 & 10.46 & 4.60 & 10.57  & 73.89 & 0.19 & 0.41 & 0.72 \\
         &p-trifocal \cite{hartley2003multiple} & 0.37 & 0.76 & 0.77 & 1.59 & 1.24 & 2.52 & 1.57 & 3.31 & 24.35 & 111.4 & 24.91 & 108.0 & 24.83 & 105.2 & 24.52 & 107.4 & \sbest0.04 & 0.08 & 0.14 & 0.22 \\
         &NEC \cite{kneip2013direct} & \tbest0.20 & \tbest0.32 & 0.37 & 0.61 & 0.55 & 0.90 & 0.71 & 1.17 & \best\textbf{0.00} & \best\textbf{0.00} & 0.33 & 0.72 & 0.57 & 1.26 & 0.79 & 1.71  & \best\textbf{0.00}  & 0.06 & 0.13 & 0.20 \\
         &PNEC \cite{muhle2022probabilistic} & \sbest0.13 & \sbest0.22 & \tbest0.27 &\tbest 0.46 & \tbest0.40 & \tbest0.69 & \tbest0.55 & \tbest0.97 & \best\textbf{0.00} & \best\textbf{0.00} & \tbest0.22 & \tbest0.52 & \tbest0.45 & \tbest1.06 & \tbest0.63 & \tbest1.41 & \best\textbf{0.00} & \sbest0.04 & \sbest0.08 & \sbest0.13  \\ 
         \midrule
         \multirow{2}{*}{\rotatebox[origin=c]{90}{line}} &l-trifocal \cite{hartley2003multiple} & 2.94 & 4.26 & 6.15 & 8.85 & 9.10 & 13.16 & 12.12 & 17.70 & 163.2 & 115.1 & 157.0 & 107.9 & 155.1 & 106.7 & 156.6 & 108.8  & 158.4 & 150.9 & 151.4 & 152.2 \\
         &NBC (Ours w/o IRLS \& Points) & 0.28 & 0.74 & 0.54 & 1.58 & 0.88 & 2.69 & 1.19 & 3.97 & \best\textbf{0.00}  & \best\textbf{0.00} & 0.41 & 1.23 &  0.86 & 2.83 & 1.32 & 4.79 & 0.58 & 0.07 & 0.13 & 0.20  \\
         \midrule
         \multirow{3}{*}{\rotatebox[origin=c]{90}{point-line}}&pl-trifocal \cite{hartley2003multiple} & 0.23 & 0.45 & 0.49 & 0.95 & 0.72 & 1.42 & 0.93 & 1.88 & 24.37 & 111.0 & 23.20 & 105.4 & 23.22 & 103.1 & 23.49 & 106.9 & \tbest0.07 & 0.07 & 0.13 & 0.21 \\
         &NBEC (Ours w/o IRLS) & \sbest0.13 & \sbest0.22 & \sbest0.25 & \sbest0.44 & \sbest0.38 & \sbest0.66 & \sbest0.49 & \sbest0.87 & \best\textbf{0.00}  & \best\textbf{0.00} & \sbest0.16 & \sbest0.36 &  \sbest0.32 & \sbest0.72 & \sbest0.51 & \sbest1.13 & \best\textbf{0.00} & \tbest0.06 & \tbest0.13 & \tbest0.19  \\
         &RT$^2$PL (Ours) & \best\textbf{0.07} & \best\textbf{0.14} & \best\textbf{0.14} & \best\textbf{0.27} & \best\textbf{0.21} & \best\textbf{0.40} & \best\textbf{0.29} & \best\textbf{0.55} & \best\textbf{0.00} & \best\textbf{0.00} & \best\textbf{0.08} & \best\textbf{0.21} & \best\textbf{0.16} & \best\textbf{0.42} & \best\textbf{0.25} &  \best\textbf{0.64} & \best\textbf{0.00} & \best\textbf{0.03} & \best\textbf{0.05} & \best\textbf{0.07}  \\
        \bottomrule[1.0pt]
    \end{tabular}
    }
    \begin{tablenotes}
    \footnotesize
        \item[$^\dagger$] \textbf{For fair comparisons, all algorithms conduct their non-minimal version.} 7pt falls back to 8pt under this configuration. 
        \item[] The experiment at each noise level is repeated 1000 times, and the average values are presented. 
        \item[] Best results are highlighted as \colorbox{Green!25}{\bf first}, \colorbox{SpringGreen!45}{second}, and \colorbox{Yellow!30}{third}.
   \end{tablenotes}
    \label{tab:major_test}
    \vspace{-0.1in}
    \end{threeparttable}
\end{table*}
\begin{figure*}[t]
\centering
     \subfigure[General case]{\includegraphics[height=0.125\linewidth]{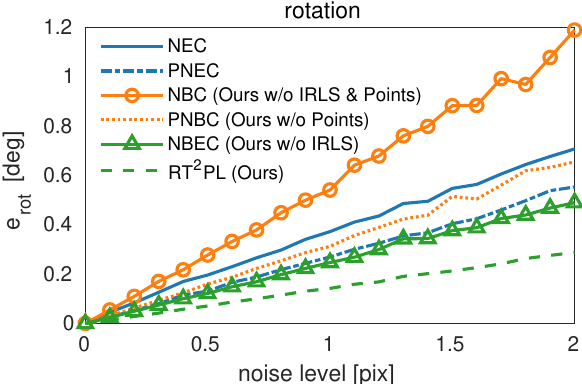}
     \includegraphics[height=0.125\linewidth]{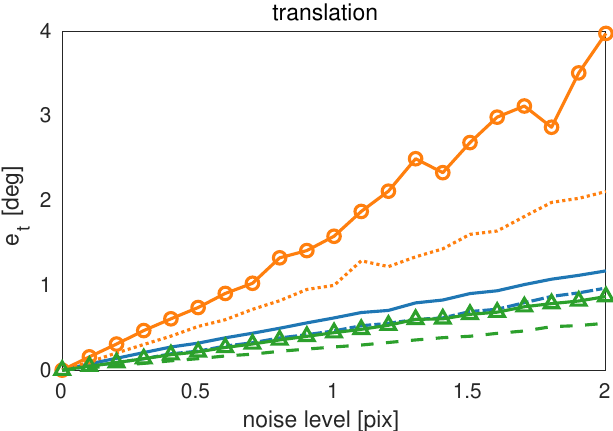}
      \label{fig:general_case_est}}
  \subfigure[Planar degeneracy]{\includegraphics[height=0.125\linewidth]{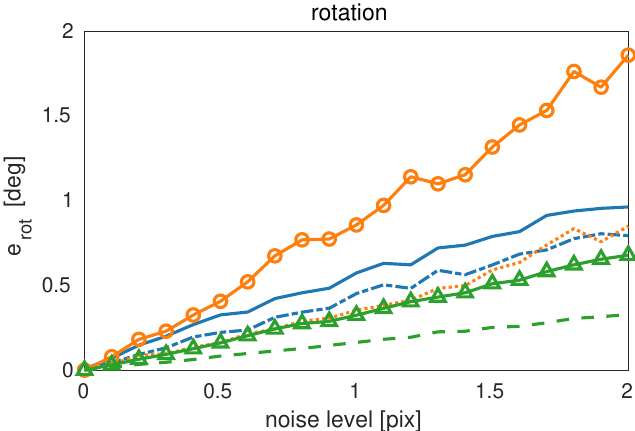}
    \includegraphics[height=0.125\linewidth]{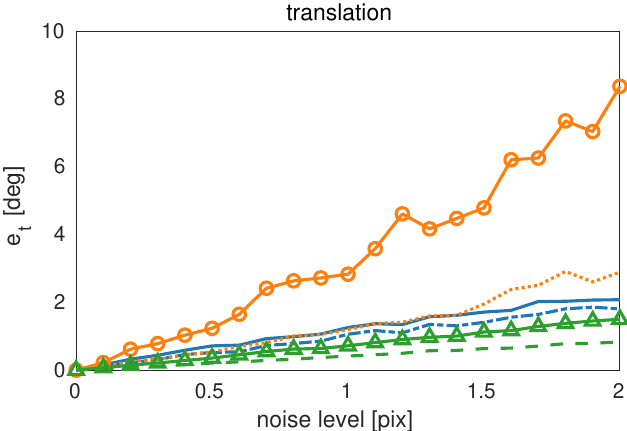}
    \label{fig:planar_test}}
     \subfigure[Pure rotation degeneracy]{\includegraphics[height=0.125\linewidth]{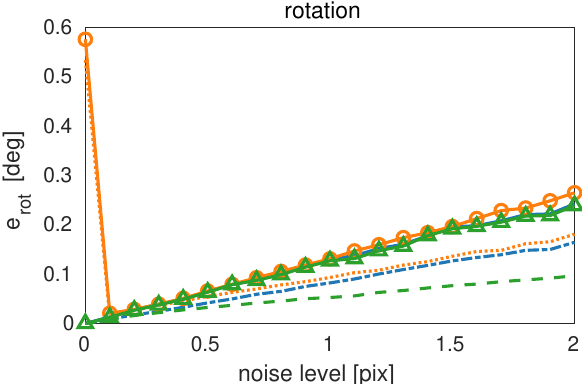}
     \label{fig:pure_rotation_test}}
   \caption{\textbf{Ablation experiments.}  The confighuration is set same as Tab \ref{tab:major_test}. Each value is averaged over 1000 random experiments.
    }
   \vspace{-0.1in}  
   \label{fig:ablation_exp}
   \vspace{-0.1in}
\end{figure*}

\subsection{Line-based LiGT}
 For lines, according to the trifocal tensor constraints \cite{hartley2003multiple}, 
 \begin{equation}
    \begin{aligned}
        \bold{0} &= [{}^0\bold{n}]_\times\left[({}^2\bold{t}_0^T{}^2\bold{n})\bold{R}_{01}{}^1\bold{n} - ({}^1\bold{t}_0^T{}^1\bold{n})\bold{R}_{02}{}^2\bold{n}\right]\\
        &= ([{}^0\bold{n}]_\times\bold{R}_{01}{}^1\bold{n}){}^2\bold{n}^T{}^2\bold{t}_0{}-([{}^0\bold{n}]_\times\bold{R}_{02}{}^2\bold{n}){}^1\bold{n}^T{}^1\bold{t}_0{},
    \end{aligned}
    \label{eq:trifocal_con}
 \end{equation}
 we proposed a line-based LiGT. Specifically, we refined the constraints Eq. (\ref{eq:trifocal_con}) in two ways. First, we represent the relative translation in the global frame. Second, it is observed that the direction of the vector $([{}^0\bold{n}]_\times\bold{R}_{01}{}^1\bold{n})$ and the vector $([{}^0\bold{n}]_\times\bold{R}_{02}{}^2\bold{n})$ are both equal to the line direction represented in the frame $0$ (${}^0\bold{r}$) ideally. To mitigate the impact of noise, we retain only the magnitudes of the two vectors and replace their directions with the estimated direction of the line. Thus the coefficient matrices are given below:
\begin{equation}
    \begin{aligned}
        \bold{B} &=  \mathrm{sgn}({}^0\bold{r}^T [{}^0\bold{n}]_\times\bold{R}_{01}{}^1\bold{n})\left\|[{}^0\bold{n}]_\times\bold{R}_{01}{}^1\bold{n}\right\|{}^2\bold{n}^T\bold{R}_{2 G}\\
        &-\mathrm{sgn}({}^0\bold{r}^T [{}^0\bold{n}]_\times\bold{R}_{02}{}^2\bold{n})\left\|[{}^0\bold{n}]_\times\bold{R}_{02}{}^2\bold{n}\right\|
        {}^1\bold{n}^T\bold{R}_{1 G}\\
        \bold{C} &= \mathrm{sgn}({}^0\bold{r}^T [{}^0\bold{n}]_\times\bold{R}_{02}{}^2\bold{n})\left\|[{}^0\bold{n}]_\times\bold{R}_{02}{}^2\bold{n}\right\|
        {}^1\bold{n}^T\bold{R}_{1 G} \\
        \bold{D} &= -(\bold{B} + \bold{C}),
        \label{eq:trifocal_LiGT}
    \end{aligned}
\end{equation}
where, $\mathrm{sgn}(\cdot)$ is a sign function.
When the minimal eigenvalue corresponding to the line direction is large, the line will be regarded as a bad line and will be removed. Therefore, rotation errors and observation noise have little impact on the value of the sign function in Eq.  (\ref{eq:trifocal_LiGT}). The degrees of the rotation and feature observation components in Eq. (\ref{eq:trifocal_LiGT}) are $2$ and $3$, which are lower than those in the LiGT form proposed by \cite{holt1994motion}, namely Holt-LiGT, which are 3 and 5, respectively. Please refer to App. \ref{app:Holt-LiGT} for the entire derivation. Besides, we also construct a line-based LiGT with a line-based depth-pose-only (DPO) constraint like the original LiGT\cite{cai2021pose}. We will also prove that the proposed line-based LiGT form (Eq. (\ref{eq:trifocal_LiGT})) outperforms the line-based LiGT from with DPO constraints in App. \ref{app:dpo_line}.


\subsection{Translation Part of RT$^2$PL}
For point-line-based LiGT, we use Eq. (\ref{eq:newligt_bcd}) for point features and Eq. (\ref{eq:trifocal_LiGT}) for line features. The Singular Value Decomposition (SVD) is used to solve the entire linear problem, ensuring efficiency and independence from the number of features.

\section{Experiments}
\label{sec:experiments}
In this section, we first conduct thorough experiments using synthetic data to validate both the fusion strategy and the uncertainty weighting strategy. Simultaneously, the experiments with synthetic data will demonstrate the advantages of our method in terms of degeneracy resilience and noise resistance. Then experiments on real-world data will further confirm the accuracy advantages of our method.
\begin{figure}[t]
\centering
    \includegraphics[width=0.98\linewidth]{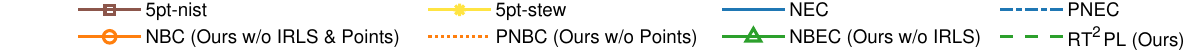}
\subfigure[rotation error]{
     \includegraphics[width=0.45\linewidth]{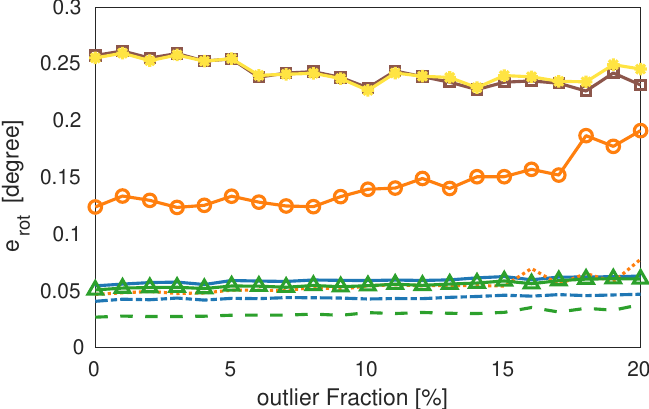}
}
\subfigure[translation error]{
     \includegraphics[width=0.45\linewidth]{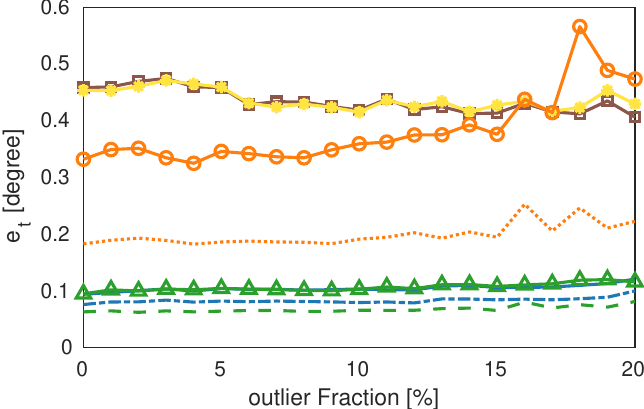}
}
   \caption{\textbf{Convergence and resilience to outliers.} For this experiment, 100 random points and 100 random lines are generated. All algorithms are embedded into a RANSAC scheme with the same outlier threshold and inlier criteria. Following the approach outlined in \cite{kneip2013direct}, five features are used as the sample set for 5pt-nist and 5pt-stew. Ten features are used as the sample sets for all non-minimal solvers.}
   \label{fig:outlier}
\end{figure}
\begin{figure}[t]
\centering
\subfigure[RANSAC test]{
     \includegraphics[width=0.45\linewidth]{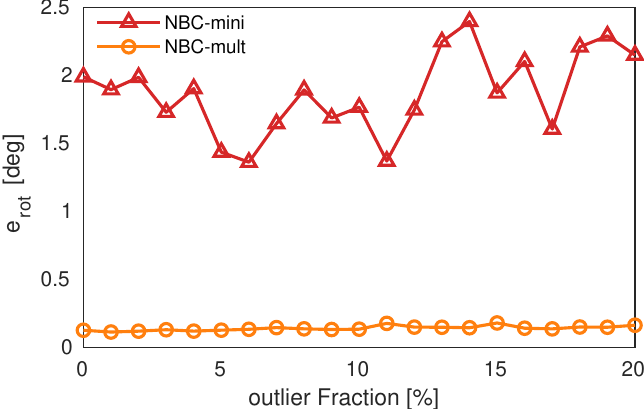}
     \label{fig:NBC_ransac}
}
\subfigure[initial value resilience test]{
     \includegraphics[width=0.45\linewidth]{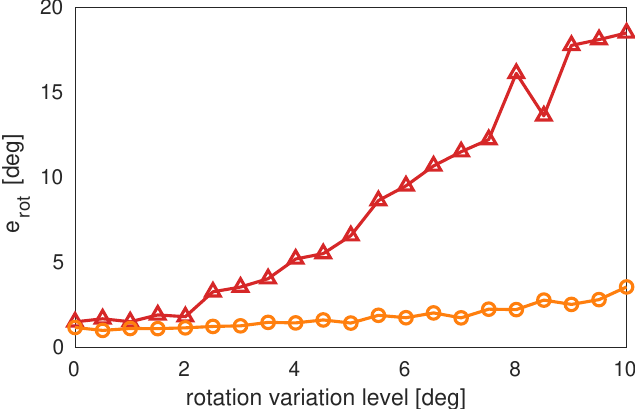}
     \label{fig:NBC_init}
}
   \caption{\textbf{Convergence analysis for two NBC forms.} (a) Results of RANSAC with random initialization. The configuration is the same as that for Fig. \ref{fig:outlier}. (b) Initial value resilience test. We set the initial values with deviations ranging from $0^\circ$ to $10^\circ$ near the true values to test the algorithm convergence without outliers. The noise is fixed to 0.5 pixels.}
   \label{fig:convergence_test}
\end{figure}
\begin{figure}[t]
\vspace{0.3in}
\centering
\subfigure[NBC-mini]{
     \includegraphics[width=0.49\linewidth]{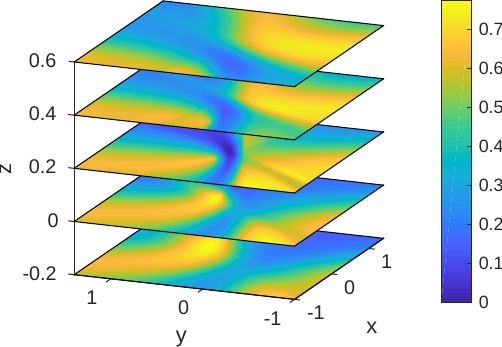}
     \includegraphics[width=0.49\linewidth]{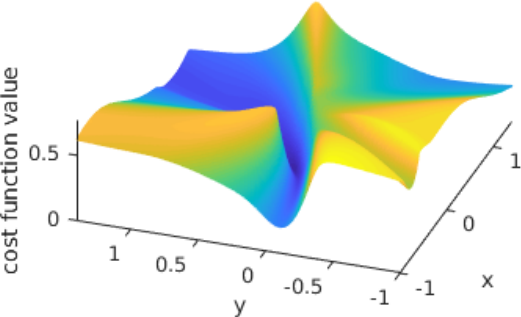}
}
\subfigure[NBC-mult]{
     \includegraphics[width=0.49\linewidth]{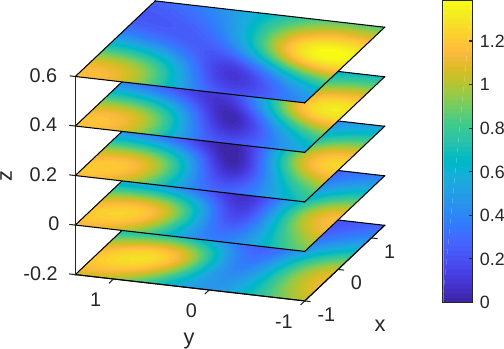}
     \includegraphics[width=0.49\linewidth]{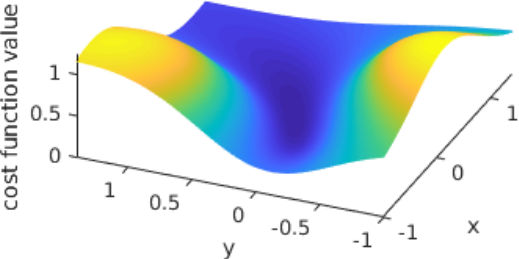}
}
   \caption{\textbf{Behavior of the NBC cost function.} The testing environment consists of 10 lines and no observation noise is added. Because the cost function involved 6 unknowns, for visualization, we set one relative rotation as the ground truth. We display the values of the cost function as the Cayley parameters of the other rotation change. These parameters take on the ground truth at $[0.2,0.2,0.2]$. The right figures display the cost function value changes when $z=0.2$. The cost function in the multiple form exhibits smoother performance. Besides, the cost function in the multiple form has a wider basin of convergence than that in the minimal form.}
   \label{fig:cost_function_behavior}
\end{figure}

\subsection{Experiments in Synthetic Scenes}
\label{sec:synthetic_exp}
To evaluate the performance of the proposed methods, we generate synthetic scenes based on the manner in \cite{kneip2013direct}. Three frames are generated, with the first frame fixed. The orientation of the second frame relative to the first is generated using random Euler angles constrained to an absolute value of $0.5$ radians. The orientation of the third frame relative to the second is generated using the same procedure. The two relative translations between consecutive frames are generated with a uniformly distributed random direction and a maximum magnitude of $2$. The 3D point landmarks and the 3D endpoints of the line landmarks are obtained from uniformly distributed random points around the origin of the first frame at distance between $4$ and $8$. Simultaneously, the focal length of the virtual camera is set to $800$. Gaussian noise is introduced to perturb each point and each endpoint in the images.
The standard deviation of the Gaussian noise serves as the noise level for both point and line observations. We compare our method on the synthetic scenes to nine classical or state-of-the-art methods: 
\begin{itemize}
    \item 7 point-based pose estimation methods: namely \textbf{5pt-nist} \cite{nister2004efficient}, \textbf{5pt-stew} \cite{stewenius2006recent}, \textbf{7pt} \cite{hartley2003multiple}, \textbf{8pt} \cite{hartley1997defense}, \textbf{NEC} \cite{kneip2013direct}, \textbf{PNEC}\cite{muhle2022probabilistic}, and a point-based trifocal tensor method (\textbf{p-trifocal} \cite{hartley2003multiple}). 
    \item 1 line-based pose estimation method: line-based trifocal tensor method (\textbf{l-trifocal} \cite{hartley2003multiple})
    \item 1 point-line-based pose estimation method: point-line-based trifocal tensor method (\textbf{pl-trifocal} \cite{hartley2003multiple}).  
\end{itemize}
The rotation estimation error metric $e_{rot}$ is defined by the sum of the angle differences of two consecutive relative rotations:
\begin{equation}
    e_{rot} \coloneqq \angle (\bold{R}_{01}^T \tilde{\bold{R}}_{01})+ \angle (\bold{R}_{12}^T \tilde{\bold{R}}_{12}),
\end{equation}
where, $\bold{R}$ denotes the ground truth and $\tilde{\bold{R}}$ denotes the estimation, The function $\angle(\cdot)$ returns the angle of the rotation matrix.
The translation estimation error metric $e_{t}$ is defined by the sum of the angle differences of two consecutive relative translation directions:
\begin{equation}
    e_{t} \coloneqq \arccos({}^0\bold{t}_{1}^{T}{}^0\tilde{\bold{t}}_{1}) +  \arccos({}^{1}\bold{t}_{2}^{T}{}^{1}\tilde{\bold{t}}_{2}),
\end{equation}
where, $\bold{t}$ denotes the ground truth  and $\tilde{\bold{t}}$ denotes the estimation.

\subsubsection{Image Noise Resilience and Ablation Test}
\label{sec:Noise_Resilience}
We generate 15 random points and 15 random lines (to compare our method with the line-based trifocal tensor method, which needs at least 13 lines, plus two more lines for trifocal tensor decomposition) without outliers. Different image noise levels, ranging from $0$ pixels to $2$ pixels, are added. The step size of the noise level is $0.1$ pixels. Each experiment with a given image noise level is repeated 1000 times.
The cost functions in our methods, as well as that in NEC \cite{kneip2013direct} and PNEC \cite{muhle2022probabilistic}, are non-convex, making them sensitive to the initial point. We set the starting value as a uniform variation around the true rotation, following the approach in \cite{kneip2013direct}, to ensure that the global minimum is identified for a proper evaluation of noise resilience. Further discussion about convergence will be provided in Sec. \ref{sec:convergence}. To demonstrate the benefits of decoupled estimation, we simulate planar degeneracy cases and pure rotation degeneracy cases by imposing features on a plane and setting translation to zero, respectively.

\begin{figure}[t]
\centering
{
     \includegraphics[width=0.98\linewidth]{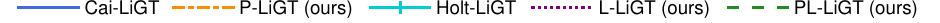}
}
{
\subfigure[rotation error resilience]{
     \includegraphics[width=0.45\linewidth]{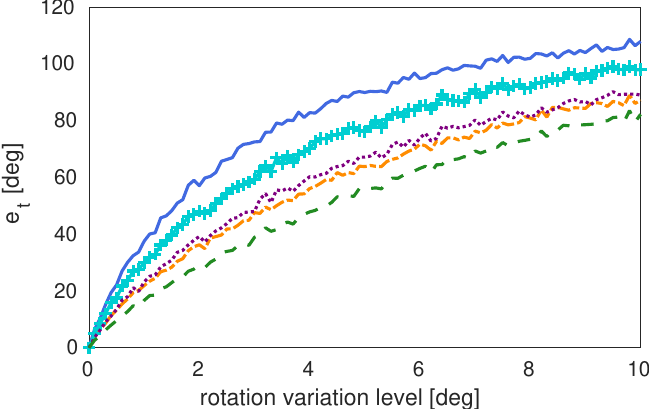}
}
\subfigure[observation noise resilience]{
     \includegraphics[width=0.45\linewidth]{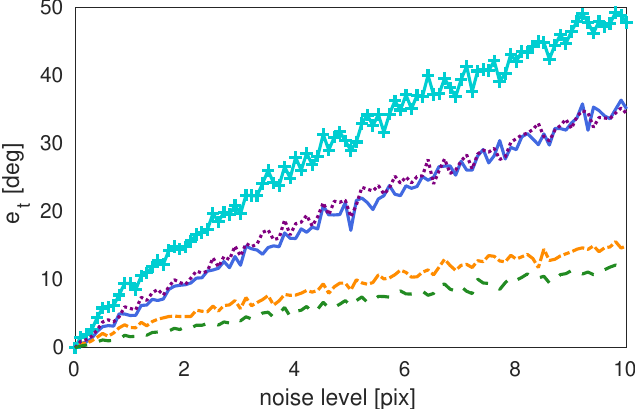}
}
}
   \caption{\textbf{Resilience analysis for LiGT.} (a) To assess rotation error resilience, the noise level is set to $0$ pixels, and rotation variation ranges from $0^\circ$ to $10^\circ$. (b) To test observation noise resilience, the rotation variation level is fixed as $0^\circ$, and the noise level varies from $0$ pixels to $10$ pixels. Each data point in the plots represents the mean of 1000 trails.}
   \label{fig:LiGT_test}
\end{figure}
\begin{table}[t]
\vspace{-0.1in}
   \centering
   \caption{Estimation time with 20\% outlier fraction (ms)}
   \begin{threeparttable}
    \resizebox{0.99\linewidth}{!}{
    \begin{tabular}{lcccccc}
    \toprule
      & 5pt-stew\cite{stewenius2006recent}& NEC\cite{muhle2022probabilistic} & PNEC \cite{muhle2022probabilistic} & NBC (Ours) & PNBC (Ours) & RT$^2$PL (Ours)\\
    \midrule
    RASANC$^\star$            & 2.44/38 & 4.53/63  & 4.53/63 & 6.70$\times$2$^\dagger$/41     & 6.70$\times$2$^\dagger$/41 & (4.53+6.70)$\times$2$^\dagger$/63+41 \\
    IRLS+Trans.$^\ddag$    & -       & -        & 1.17    & -                    &0.46$\times$2$^\dagger$     & 1.06$\times$2$^\dagger$ \\
    Per frame Est.          & 2.44    & 4.53     & 5.70    & 6.70                 &7.16              & 12.29 \\
    \bottomrule
    \end{tabular}
    }
    \begin{tablenotes}[flushleft]
    \vspace{0.1in}
    \footnotesize
        \item[$^\star$] Present the time of RANSAC and the number of iterations.
        \item[$^\dagger$] $\times2$ means that those methods deal with two relative poses together. 
        \item[$^\ddag$] For NEC and NBC, the translation is computed in the RASANC frame-\\work. For PNEC, we need the translation for weighting so that the transl-\\ation is computed in each loop of IRLS. For lines, we do not need the \\translation in the loop, therefore, the translation is computed after IRLS.
   \end{tablenotes}
    \label{tab:running_time}
    \end{threeparttable}
\end{table}

The results in Tab. \ref{tab:major_test} show that the proposed method (RT$^2$PL) outperforms all of the comparative methods, and its advantage becomes more pronounced as the noise level increases. This success is mainly attributed to the incorporation of uncertainty weighting in the cost function and the fusion of points and lines, as evidenced by the results of the ablation experiment shown in Fig. \ref{fig:ablation_exp}. Furthermore, the results demonstrate that methods employing a decoupled estimation scheme (NEC, PNEC, NBEC, RT$^2$PL) significantly outperform those with mixture parameter problems when degeneracy occurs. It is worth noting that the estimation error with NBC constraints, which also utilizes the decoupled scheme, is not zero when the noise level equals zero in pure rotation cases, as shown in the second-to-last row of Tab. \ref{tab:major_test} and Fig. \ref{fig:pure_rotation_test}. This indicates that NBC cannot handle pure rotation degeneracy cases. It is because when two arbitrary back-projected planes overlap, the position of the third back-projected plane will be unconstrained, i.e., the normals of the back-projected planes of an arbitrary line feature are always coplanar. When the noise level is not zero, the back-projected plane will not strictly overlap which leads to smaller estimation errors. Although NBC is degenerate in pure rotation cases, it can still provide some useful constraints for pose estimation. Thanks to the combination of point and line constraints, the proposed method (RT$^2$PL) achieves a significant improvement in accuracy compared to PNEC \cite{muhle2022probabilistic} in pure rotation cases.

\begin{figure}
    \centering
    \includegraphics[height=0.45\linewidth]{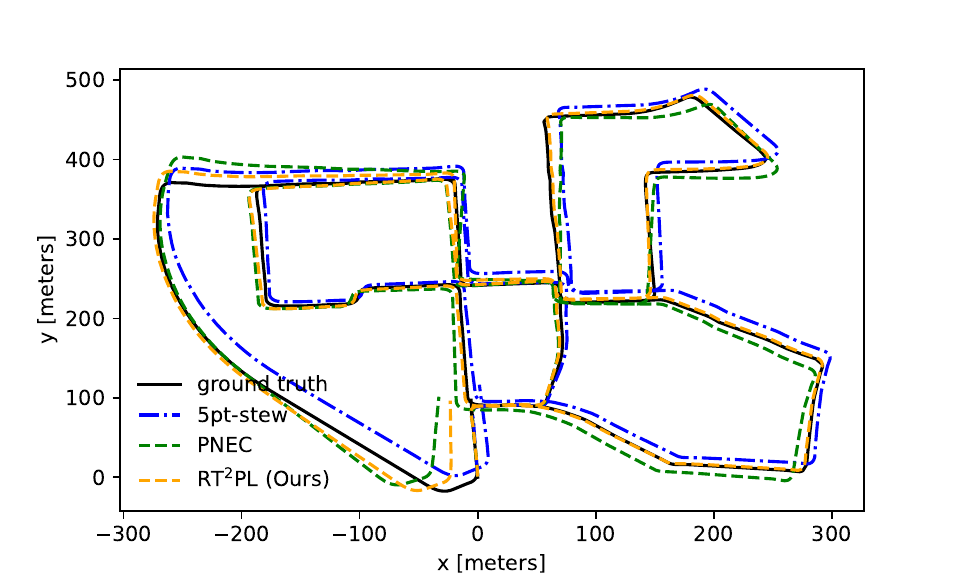}
    \caption{Trajector on KITTI Seq. 00}
    \label{fig:kitti_00}
\end{figure}

\begin{table}[t]
    \centering
    \caption{Comparison on EuRoC datasets}
    \resizebox{0.96\linewidth}{!}{
    \begin{tabular}{lccc|ccc|ccc}
    \toprule[1.0pt]
    & \multicolumn{3}{c}{5pt-stew\cite{stewenius2006recent}} & \multicolumn{3}{c}{PNEC\cite{muhle2022probabilistic}} &  \multicolumn{3}{c}{RT$^2$PL (Ours)} \\
    &$\rm{RRE}$&$\rm{RTE}$&$\rm{NUM}$&$\rm{RRE}$&$\rm{RTE}$&$\rm{NUM}$&$\rm{RRE}$&$\rm{RTE}$&$\rm{NUM}$\\ 
    Seq.&$(^\circ)\downarrow$&$(^\circ)\downarrow$&$\uparrow$&$(^\circ)\downarrow$&$(^\circ)\downarrow$&$\uparrow$&$(^\circ)\downarrow$&$(^\circ)\downarrow$&$\uparrow$\\
    \midrule
        00 & \tbest0.111 & \tbest3.397 & \tbest3511 & \sbest0.086 & \sbest2.255 & \sbest4124 & \best0.085 & \best2.241 & \best4133 \\
        01 & \tbest0.798 & \tbest33.254 & \tbest340 & \sbest0.102 & \sbest10.328 & \sbest479 & \best0.100 & \best10.037 & \best487 \\
        02 & \tbest0.098 & \tbest1.942 & \tbest3465 & \sbest0.060 & \sbest1.528 & \sbest4166 & \best0.059 & \best1.525 & \best4186 \\
        03 & \tbest0.072 & \tbest3.731 & \tbest657 & \sbest0.044 & \sbest1.728 & \sbest757 & \best0.043 & \sbest1.735 & \best758 \\
        04 & \tbest0.080 & \tbest0.924 & \tbest66 & \sbest0.035 & \sbest0.585 & \sbest178 & \best0.033 & \best0.565 & \best184 \\
        05 & \tbest0.076 & \tbest4.131 & \tbest1891 & \sbest0.048 & \sbest1.721 & \sbest2343 & \best0.047 &\best 1.644 & \best2361 \\
        06 & \tbest0.070 & \tbest1.288 & \tbest506 & \sbest0.034 & \sbest0.919 & \sbest817 & \best0.033 & \best0.907 & \best840 \\
        07 & \tbest0.132 & \tbest11.097 & \tbest728 & \best0.039 & \sbest4.561 & \best924 & \best0.039 & \best4.508 & \best924 \\
        08 & \tbest0.080 & \tbest8.138 & \tbest2815 & \sbest0.044 & \best7.526 & \sbest3560 & \best0.043 & \sbest7.533 & \best3575 \\
        09 & \tbest0.083 & \tbest1.367 & \tbest1258 & \sbest0.046 & \sbest1.039 & \sbest1473 & \best0.045 & \best1.031 & \best1477 \\
        10 & \tbest0.092 & \tbest6.735 & \tbest874 & \sbest0.051 & \sbest2.457 & \sbest1048 &\best 0.050 & \best2.432 & \best1053 \\
    \midrule
     Avg.& \tbest0.154 & \tbest6.909 & \tbest1465 & \sbest0.054 & \sbest3.150 & \sbest1806 & \best0.052 & \best3.105 & \best1816 \\  
    \bottomrule
    \end{tabular}
    }
    \label{tab:kitti_results}
\end{table}

\subsubsection{Convergence and Resilience to Outliers}
\label{sec:convergence}
We employed the RANSAC scheme \cite{fischler1981random} to enhance the robustness of the proposed algorithms in handling outliers. It is compared with the five-point method \cite{nister2004efficient, stewenius2006recent}, NEC \cite{kneip2013direct}, and PNEC \cite{muhle2022probabilistic} in a synthetic scene containing 100 points and 100 lines, with up to $20\%$ outliers in each type of measurement. The noise is fixed to 0.5 pixels following the approach in \cite{kneip2013direct}. For non-convex methods (NEC, PNEC, and all proposed methods), we add random variation in each RANSAC iteration,  similar to the approach in \cite{kneip2013direct}, to try to avoid local minima. The initial value is set to a zero vector. In the case of IRLS-based methods (PNEC, PNBC, RT$^2$PL), the RANSAC scheme is only conducted during the initial loop of IRLS. The results in Fig. \ref{fig:outlier} show that the random variation scheme is also valid for the proposed line-based rotation estimation. Due to the higher sensitivity of line features to noise, as observed in Fig. \ref{fig:ablation_exp} and Fig. \ref{fig:outlier}, after executing separate RANSAC operations for points and lines for RT$^2$PL, we prefer using the initial estimation generated by point-based RANSAC with random variation when a sufficient number of point features is available. Then the inliers of points and lines will be used together to produce the initial estimation for the IRLS process. 

As mentioned in Sec. \ref{sec:NBC}, we propose two forms of NBC constraints i.e., NBC-mini and NBC-mult. We adopt the NBC-mult as the constraint for the line part in RT$^2$PL.This choice is based on the better behavior of the cost function with NBC-mult constraint, as outlined in Fig. \ref{fig:cost_function_behavior}. The results of a further test for convergence of the two forms are shown in Fig. \ref{fig:convergence_test}. Compared with the NBC-mini form, the NBC-mult form has better resilience to initial value as illustrated in Fig. \ref{fig:NBC_init}. Therefore, as shown in Fig \ref{fig:NBC_ransac}, NBC-mult will provide more accurate results within the RANSAC framework with the random variation scheme.

\begin{figure}
    \centering
    \includegraphics[height=0.45\linewidth]{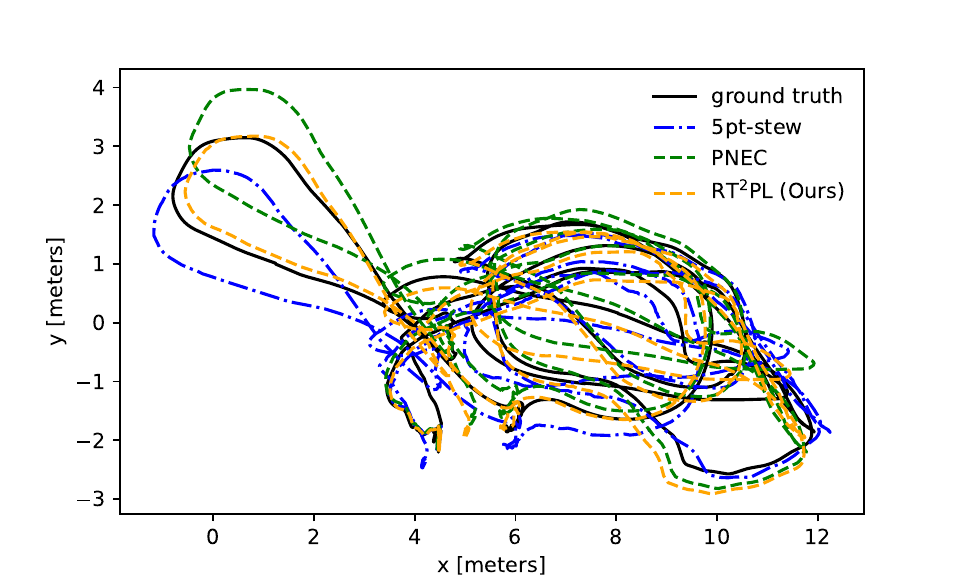}
    \caption{Trajectory on EuRoC Seq. MH\_03}
    \label{fig:mh_03}
\end{figure}

\begin{table}[t]
    \centering
    \caption{Comparison on EuRoC datasets}
    \resizebox{0.96\linewidth}{!}{
    \begin{tabular}{lccc|ccc|ccc}
    \toprule[1.0pt]
    & \multicolumn{3}{c}{5pt-stew\cite{stewenius2006recent}} & \multicolumn{3}{c}{PNEC\cite{muhle2022probabilistic}} &  \multicolumn{3}{c}{RT$^2$PL (Ours)} \\
    &$\rm{RRE}$&$\rm{RTE}$&$\rm{NUM}$&$\rm{RRE}$&$\rm{RTE}$&$\rm{NUM}$&$\rm{RRE}$&$\rm{RTE}$&$\rm{NUM}$\\ 
    Seq.&$(^\circ)\downarrow$&$(^\circ)\downarrow$&$\uparrow$&$(^\circ)\downarrow$&$(^\circ)\downarrow$&$\uparrow$&$(^\circ)\downarrow$&$(^\circ)\downarrow$&$\uparrow$\\
    \midrule
            MH\_01 & \tbest0.345 & \tbest8.463  & \tbest582 & \sbest0.174 & \sbest8.461  & \sbest621 & \best\textbf{0.171} & \best\textbf{7.172}  & \best\textbf{622}           \\
            MH\_02 & \tbest0.848 & \tbest9.496  & \tbest505 & \sbest0.202 & \sbest8.261  & \best\textbf{541} & \best\textbf{0.201} & \best\textbf{7.918}  & \best\textbf{541}           \\
            MH\_03 & \tbest0.426 & \tbest8.206  & \tbest449 & \sbest0.268 & \sbest8.251  & \best\textbf{475} & \best\textbf{0.264} & \best\textbf{8.128}  & \best475           \\
            MH\_04 & \tbest0.403 & \tbest13.150 & \tbest304 & \sbest0.227 & \sbest12.713 & \best\textbf{333} & \best\textbf{0.224} & \best\textbf{12.005} & \best\textbf{333}           \\
            MH\_05 & \tbest0.383 & \tbest12.027 & \tbest348 &\best\textbf{ 0.172} & \sbest11.892 & \best\textbf{382} & \best\textbf{0.172} & \best\textbf{11.488} & \best\textbf{382}           \\
            V1\_01 & \tbest0.392 & \tbest11.707 & \tbest512 & \sbest0.283 & \best\textbf{11.073} & \sbest536 & \best\textbf{0.282} & \sbest11.149 & \best\textbf{537}           \\
            V1\_02 & \tbest1.192 & \tbest12.201 & \tbest303 & \sbest0.723 & \best\textbf{10.291} & \sbest308 & \best\textbf{0.719} & \sbest10.335 & \best\textbf{309}           \\
            V1\_03 & \tbest2.515 & \best20.819 & \tbest336 & \sbest1.556 & \sbest21.381 & \sbest352 & \best\textbf{1.555} & \tbest21.533 & \best\textbf{353}           \\
            V2\_01 & \tbest0.587 & \tbest11.401 & \tbest397 & \best\textbf{0.315} & \best\textbf{10.177} & \sbest405 & \sbest0.316 & \sbest10.215 & \best\textbf{407}           \\
            V2\_02 & \tbest1.266 & \tbest16.413 & \tbest419 & \sbest0.898 & \best\textbf{15.443} & \sbest427 & \best\textbf{0.860} & \sbest15.607 & \best\textbf{428}           \\
            V2\_03 & \tbest3.120 & \best\textbf{28.567} & \tbest259 & \sbest1.861 & \tbest29.366 & \sbest269 & \best\textbf{1.821} & \sbest28.993 & \best\textbf{271}          \\
    \midrule
     Avg. &\tbest1.043	&\tbest13.859	&\tbest401	&\sbest0.607	&\sbest13.392	&\sbest423	&\best\textbf{0.599}	&\best\textbf{13.140}	&\best\textbf{423}\\
    \bottomrule
    \end{tabular}
    }
    \label{tab:euroc_test}
\end{table}

\subsubsection{Comparison for Translation Estimation}
\label{sec:ligt_synthic}
To assess the robustness of the proposed LiGT methods, experiments are conducted in scenes with $10$ points and $10$ lines. We take the rotation perturbation of ground truth and noisy observation as the input. The results are shown in Fig. \ref{fig:LiGT_test}. The original point-based LiGT \cite{cai2021pose} and the proposed point-based LiGT are named \textbf{Cai-LiGT} and \textbf{P-LiGT}, respectively. Line-based LiGT methods corresponding to \cite{holt1994motion} and Eq. (\ref{eq:trifocal_LiGT}) are named \textbf{Holt-LiGT} and \textbf{L-LiGT}, respectively. The proposed point-line-based LiGT is named \textbf{PL-LiGT}. As shown in Fig. \ref{fig:LiGT_test}, we can observe that the proposed PL-LiGT exhibits superior performance in both rotation error resilience and observation noise resilience tests. This primarily benefits from the fusion of point and line information and both constraints built by the minimal-degree LiGT. Concurrently, the results reveal that P-LiGT outperforms Cai-LiGT due to the lower degree of rotation and observation. The comparison among line-based LiGT methods yields consistent results. However, the degree metrics become invalid when comparing constraints with different features. For instance, L-LiGT possesses a lower observation components degree than Cai-LiGT, yet both exhibit the equivalent performance in noise resilience. This observation confirms that the line constraints are more sensitive to the observation noise than point constraints. 
\subsubsection{Runtime Analysis}
To evaluate the running time of the proposed method, we generated a synthetic scene consisting of 100 points and 100 lines, with $20\%$ outliers. The experiments were conducted on an Intel Core i9-13900KS CPU with 32GB of RAM. The results are presented in Table \ref{tab:running_time}. While RT$^2$PL takes 12.29 ms for each relative pose estimation, slightly slower than PNEC \cite{muhle2022probabilistic}, the optimization process runs in real time and proves to be more efficient than the method proposed in \cite{fabbri2020trplp}, which requires 660 ms for each step of the minimal solver.

\subsection{Experiments in Real-world Scenes}
In this section, we evaluate the proposed method on real-world data, including an outdoor dataset (KITTI dataset \cite{geiger2012we}) and two indoor datasets (EuRoC MAV dataset \cite{burri2016euroc} and CID-SIMS dataset \cite{zhang2023cid}). Before presenting the results, we will first introduce the implementation details of the odometry system.

\begin{figure}
    \centering
    \includegraphics[height=0.45\linewidth]{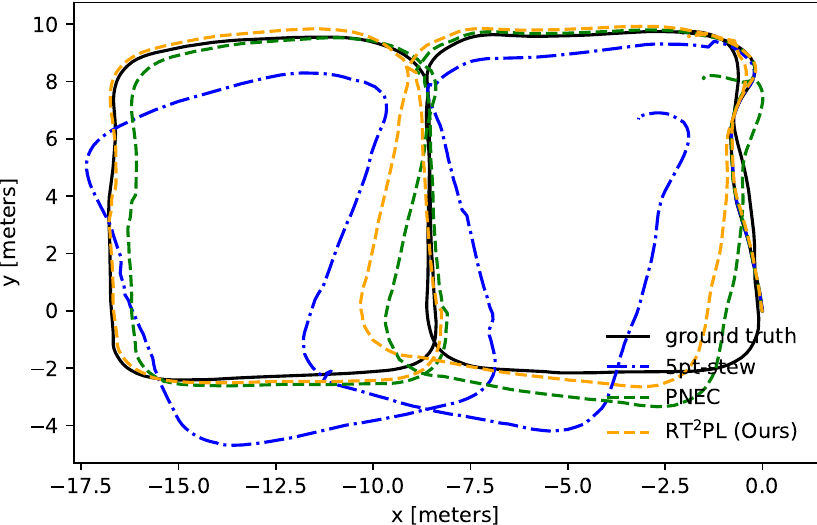}
    \caption{Trajectory on EuRoC Seq. floor3\_1}
    \label{fig:floor3_1}
\end{figure}
\begin{table}[t]
    \centering
    \caption{Comparison on CID-SIMS datasets}
    \resizebox{0.96\linewidth}{!}{
    \begin{tabular}{lccc|ccc|ccc}
    \toprule[1.0pt]
    & \multicolumn{3}{c}{5pt-stew\cite{stewenius2006recent}} & \multicolumn{3}{c}{PNEC\cite{muhle2022probabilistic}} &  \multicolumn{3}{c}{RT$^2$PL (Ours)} \\
    &$\rm{RRE}$&$\rm{RTE}$&$\rm{NUM}$&$\rm{RRE}$&$\rm{RTE}$&$\rm{NUM}$&$\rm{RRE}$&$\rm{RTE}$&$\rm{NUM}$\\ 
    Seq.&$(^\circ)\downarrow$&$(^\circ)\downarrow$&$\uparrow$&$(^\circ)\downarrow$&$(^\circ)\downarrow$&$\uparrow$&$(^\circ)\downarrow$&$(^\circ)\downarrow$&$\uparrow$\\
    \midrule
        14-13-12      & \tbest3.040 & \tbest23.500 & \tbest131  & \sbest1.338 & \best{19.479} & \sbest153  & \best{0.965} & \sbest21.148 & \best{156}   \\
        14-13-14      & \tbest1.494 & \tbest10.729 & \tbest1037 & \sbest0.722 & \best{8.301}  & \sbest1128 & \best{0.499} & \sbest8.326  & \best{1139}  \\
        apartment1\_1 & \tbest0.513 & \best{10.805} & \tbest  910  & \sbest0.237 & \tbest12.222 & \best{962}  & \best{0.235} & \sbest11.500 & \best{962}   \\
        apartment1\_2 & \tbest3.729 & \tbest13.991 & \tbest  560  & \sbest0.916 & \best{10.528} & \sbest589  & \best{0.771} & \sbest10.654 & \best{593}   \\
        apartment1\_3 & \tbest2.231 & \tbest11.296 & \tbest1051 & \sbest0.933 & \sbest11.085  & \sbest1112 & \best0.931 & \best11.051 &  \best1116  \\
        apartment2\_1 & \tbest2.337 & \tbest14.777 & \tbest 644  & \sbest0.828 & \sbest11.778 & \best{728}  & \best{0.706} & \best{11.750} & \best{728}   \\
        apartment2\_2 & \tbest1.413 & \tbest 11.437 & \tbest 1023 & \sbest0.572 & \sbest10.691 & \sbest1098 & \best{0.562} & \best{10.524} & \best{1099}  \\
        apartment2\_3 & \tbest1.265  & \tbest10.797 & \tbest1174 & \sbest0.485 & \sbest10.185 & \sbest1206 & \best0.485  & \best9.734  & \best1203   \\
        apartment3\_1 & \tbest0.284 & \tbest 12.151 & \tbest 1081 & \sbest0.228 & \tbest12.731 & \sbest1181 & \best{0.225} & \best{12.221} & \best{1187}  \\
        apartment3\_2 & \tbest4.135 & \tbest 14.633 & \tbest 660  & \sbest0.918 & \sbest13.820 & \sbest755  & \best{0.854} & \sbest14.012 & \best{756}   \\
        apartment3\_3 & \tbest2.407 & \tbest 12.068 & \tbest 949  & \sbest0.974 & \best{11.102} & \best{1057} & \best{0.828} & \sbest11.479 & \sbest1055  \\
        floor3\_1     & \tbest1.624 & \tbest 13.382 & \tbest 302  & \sbest{0.384} & \sbest7.720  & \sbest342  & \best0.378 & \best{6.978}  & \best{344}   \\
        floor3\_2     & \tbest2.336 & \tbest 11.323 & \tbest 489  & \sbest0.486 & \sbest7.596  & \sbest532  & \best{0.361} & \best{6.383}  & \best{541}   \\
        floor3\_3     & \tbest2.009 & \tbest 9.442  & \tbest 660  & \sbest0.551 & \sbest9.180  & \sbest732  & \best{0.549} & \best{8.064}  & \best{733}   \\
        floor13\_1    & \tbest5.060 & \tbest 14.697 & \tbest 381  & \sbest1.032 & \sbest8.376  & \sbest426  & \best{0.724} & \best{6.558}  & \best{430}   \\
        floor13\_2    & \tbest3.102 & \tbest 16.109 & \tbest 451  & \sbest1.080 & \sbest10.640 & \best{515}  & \best{0.719} & \best{8.890}  & \best{515}   \\
        floor14\_1    & \tbest1.001 & \tbest 5.972  & \tbest 533  & \sbest0.463 & \tbest6.714  & \sbest574  & \best{0.206} & \best{4.198}  & \best{579}   \\
        floor14\_2    & \tbest5.761 & \tbest 19.504 & \tbest 402  & \sbest3.044 & \sbest14.603 & \sbest452  & \best{0.660} & \best{10.130} & \best{462}   \\
        floor14\_3    & \tbest1.282 & \tbest 9.098  & \tbest 815  & \sbest0.518 & \sbest8.141  & \sbest885  & \best{0.438} & \best{7.910}  & \best{889}   \\
        office\_1     & \tbest0.608 & \tbest 6.899  & \tbest 448  & \sbest0.344 & \best{6.007}  & \best{466}  & \best{0.340} & \sbest6.053  & \sbest464   \\
        office\_2     & \tbest1.601 & \tbest 11.150 & \tbest 732  & \sbest0.890 & \sbest9.250  & \best{774}  & \best{0.883} & \best{9.055}  & \sbest772   \\
        office\_3     & \tbest0.965 & \tbest 9.809  & \tbest 677  & \sbest0.708 & \sbest9.521  & \sbest700  & \best{0.705} & \best{9.448}  & \best{703}   \\
    \midrule
     Avg. &\tbest2.191 &\tbest12.435 &\tbest687 &\sbest0.802	&\sbest10.440 &\sbest744 &\best0.592 &\best9.821 &\best747\\
    \bottomrule
    \end{tabular}
    }
    \label{tab:cidsims_test}
\end{table}

Our system is built upon the odometry system proposed in PNEC \cite{muhle2022probabilistic}. FAST \cite{rosten2006machine} and KLT algorithm \cite{lucas1981iterative} are leveraged for point features extraction and tracking. Compared to PNEC, our constructed odometry system has the following differences. First, to evaluate the proposed pose estimator, we extend the size of the sliding window from 2 to 3. Second, we only compare the accuracy of keyframe poses. In practical applications, keyframes in odometry are often selected based on the average disparity of points and the average included angle between the back-projected planes of the lines \cite{campos2021orb}. To ensure fair comparison, i.e., to ensure that both point-based odometry and point-line-based odometry select the same keyframes, we select keyframes at different fixed frequencies for different datasets. Specifically, for the KITTI dataset, similar to PNEC \cite{muhle2022probabilistic}, we select all frames as keyframes. For the EuRoC dataset, we select one keyframe every five frames, as done in \cite{he2023rotation}. For the CID-SIMS dataset, we select one keyframe every eight frames. The configuration for keyframe selection in different datasets is related to the frame rate and the motion speed of their data acquisition devices. Third, the LSD detector \cite{von2012lsd} and LBD algorithm \cite{zhang2013efficient} are used for line extraction and matching. For efficiency, line segments are extracted only on keyframes. For our point-line-based method, after RANSAC for points and lines, point inliers and line inliers are leveraged together to obtain the estimation for the first loop of the IRLS solver.

\begin{figure}
\centering
\subfigure[LiGT Test on KITTI Datasets]{
  \includegraphics[width=0.95\linewidth]{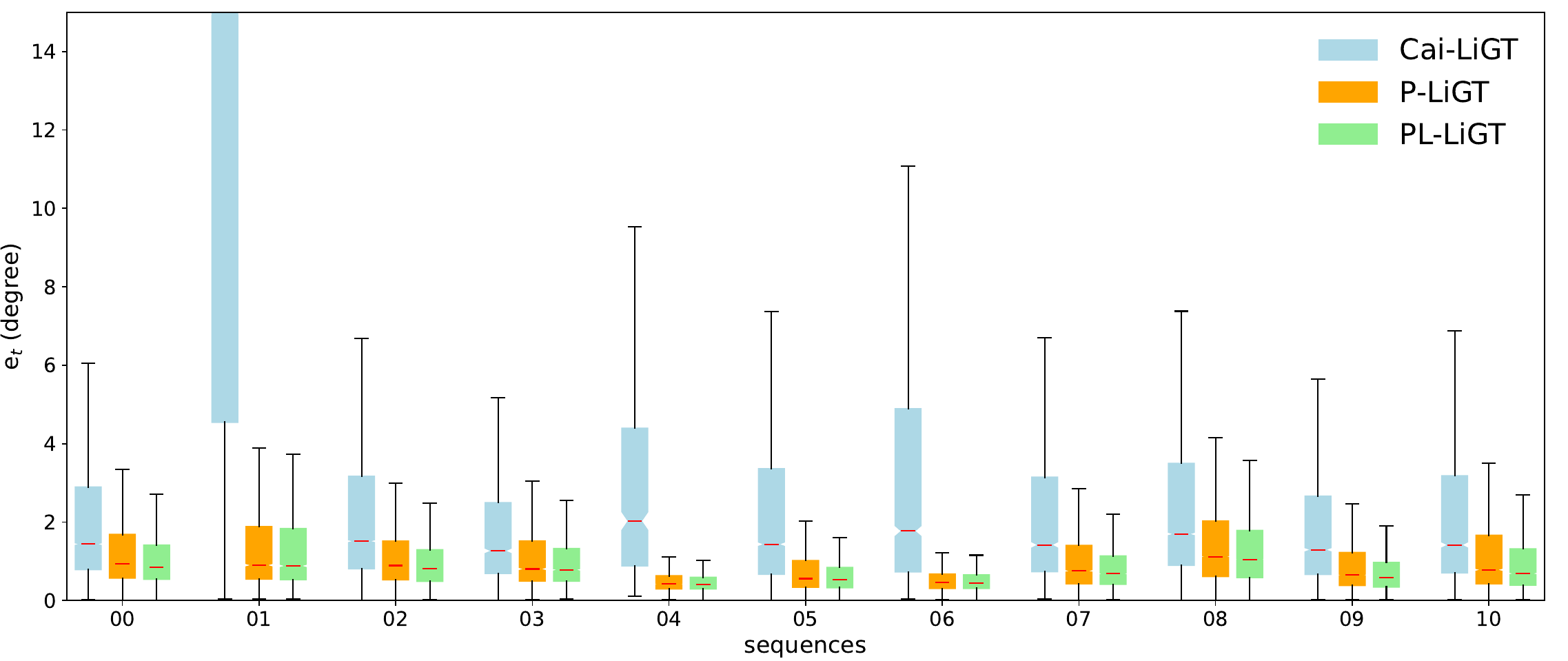}
  \label{fig:ligt_kitti}
}
\subfigure[LiGT Test on EuRoC Datasets]{
  \includegraphics[width=0.95\linewidth]{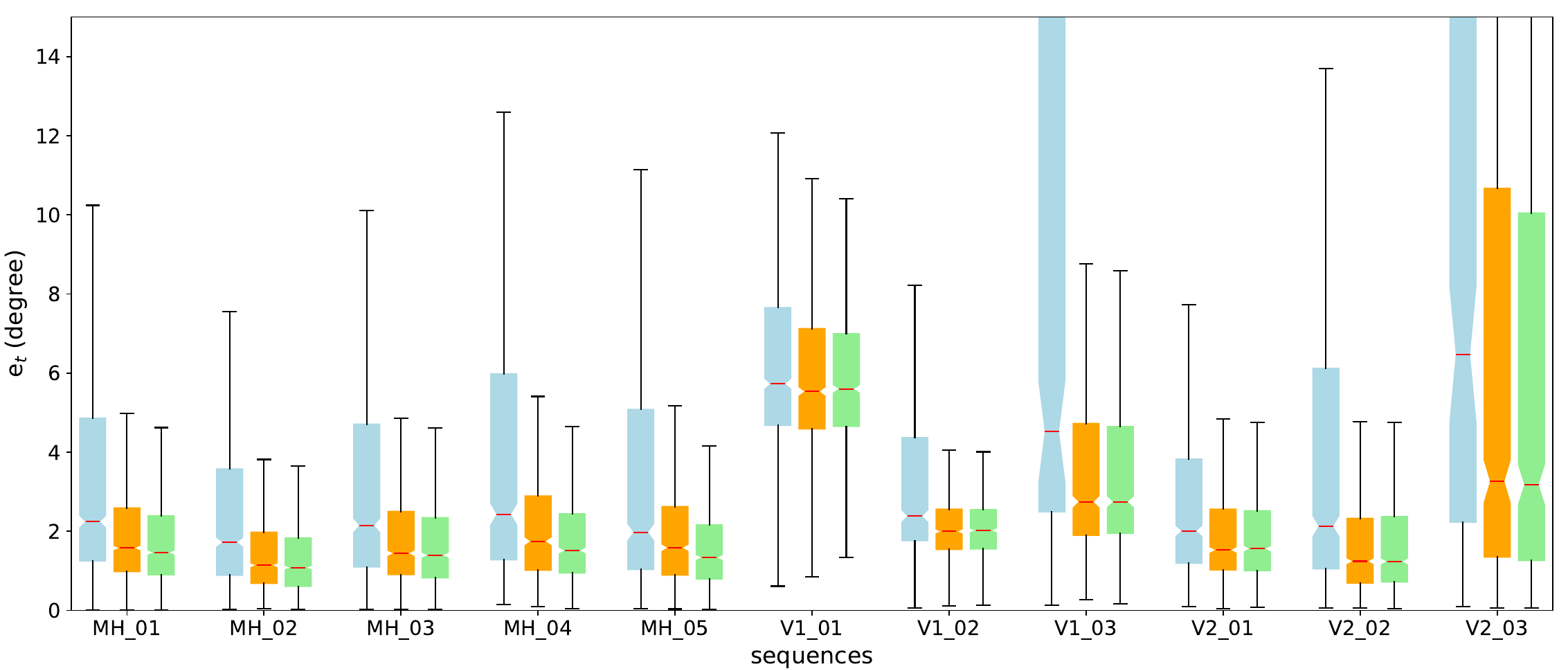}
  \label{fig:ligt_euroc}
}
\subfigure[LiGT Test on CID-SIMS Datasets]{
\centering
\begin{minipage}{0.95\linewidth}
\centering
      \includegraphics[width=1.0\linewidth]{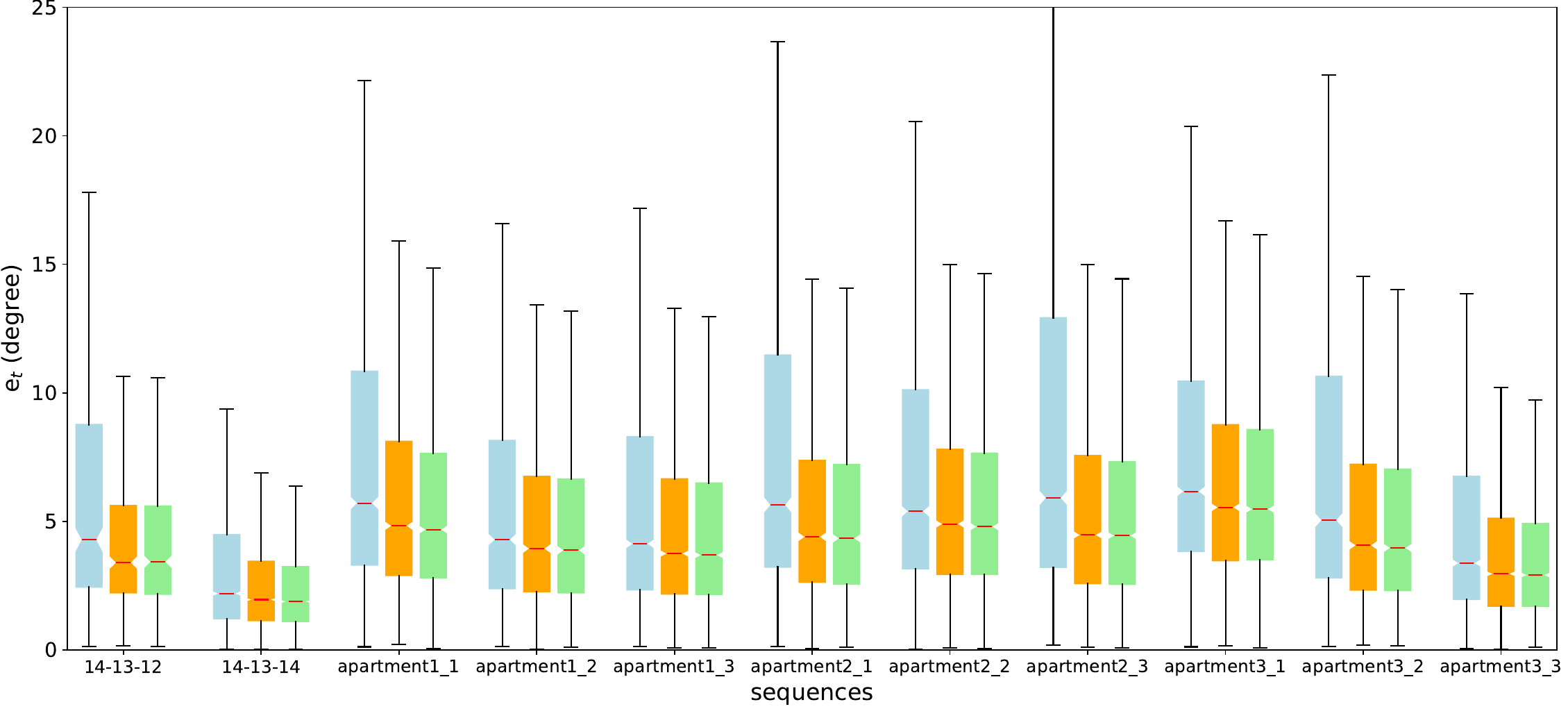}
     \includegraphics[width=1.0\linewidth] 
     {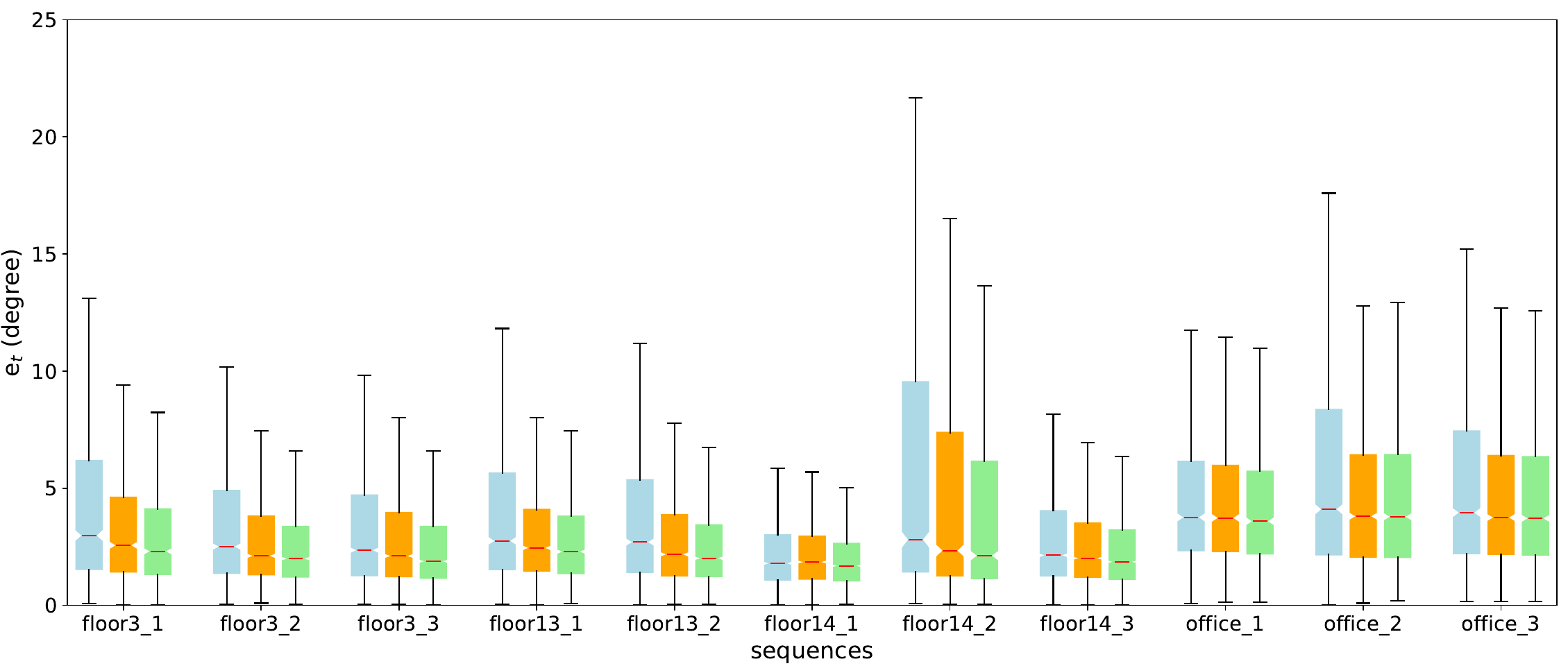}
     \vspace{0.05in}
\end{minipage}

  \label{fig:ligt_cidsims}
}
\caption{\textbf{LiGT test on real-world datasets}: Almost on all sequences of the three public datasets, P-LiGT outperforms Cai-LiGT \cite{cai2021pose}. MeanWhile, PL-LiGT gets clear improvement on some sequences compared with P-LiGT.}
\label{fig:LiGT_on_Datasets}
\end{figure}

We compare our approach with the five-point method \cite{stewenius2006recent} and PNEC \cite{muhle2022probabilistic}. All these methods are embedded into our odometry system. For our approach and PNEC, the number of IRLS iterations is set to 5. The observation noise for features is set to $1$ pixel. Additionally, we use the same magnitude of variation, set to 0.1, for our approach and PNEC in each iteration of RANSAC. We measure the accuracy of rotation and translation estimations using the root mean square errors of the \textit{relative rotation error} (RRE) and the \textit{relative translation direction error} (RTE), respectively. Furthermore, the \textit{number of successfully estimated keyframes} (NUM) is reported to compare robustness. If the $k$-th keyframe meets the condition:
\begin{equation}
\frac{\angle(\bold{R}_{i,i+1}^{T}\tilde{\bold{R}}_{i,i+1})}{\angle\bold{R}_{i,i+1}} < 0.5, \quad i = k-2, k-1,  
\end{equation}
then we define this keyframe as successfully estimated. For all methods, we compute the RRE and RTE on the successfully estimated keyframes of the method with the largest NUM for comparison.

\subsubsection{Comparison on KITTI Dataset} The KITTI dataset was captured by a car driving through a mid-sized city \cite{geiger2012we}. The dataset contains a significant amount of linear motion, with features mainly concentrated on the sides of the road.
In Tab. \ref{tab:kitti_results}, the middle performance over 5 runs of all approaches is reported. RT$^2$PL outperforms PNEC on 9 out of 11 sequences in rotation and translation accuracy. Besides, in almost all sequences, the number of successfully estimated keyframes are improved. We attribute this success to the utilization of lines and the proper fusion with points. 

Fig. \ref{fig:kitti_00} shows trajectories of RT$^2$PL, PNEC \cite{muhle2022probabilistic}, and 5pt-stew \cite{stewenius2006recent}, recovered from the estimated rotation and the estimated translation with true translation scales of all keyframes. The trajectories are aligned with the first pose. 

\subsubsection{Comparison on EuRoC MAV Dataset} The EuRoC MAV Dataset \cite{burri2016euroc} was collected on-board a micro aerial vehicle in a machine hall (MH) and a Vicon room (V).
We employ the same metrics to assess the performance of these methods on EuRoC datasets, as shown in Tab. \ref{tab:euroc_test} for a quantitative comparison. The reconstructed trajectories tested on Seq. MH\_03 are depicted in Fig. \ref{fig:mh_03}. RT$^2$PL notably enhances estimation accuracy in the sequences of the \textit{MH} series, which contain abundant high-quality lines. However, the incorporation of line features results in a reduction in pose estimation accuracy in the \textit{V} series sequences, primarily due to poor line detection caused by motion blur. Nevertheless, the estimated accuracy has not significantly declined, and the overall performance across the entire dataset has improved. 

\subsubsection{Comparison on CID-SIMS Dataset} The CID-SIMS dataset contains 22 sequences, which are mainly from a ground wheeled robot, except for two sequences that contain handheld situations when going downstairs \cite{zhang2023cid}. The dataset encompasses a variety of scenes with weak textures, as illustrated in Fig. \ref{fig:special_cases}. The results are presented in Tab. \ref{tab:cidsims_test}. From the perspective of the average performance, the proposed methods demonstrate clearer improvements on the CID-SIMS dataset compared to PNEC \cite{muhle2022probabilistic} compared with their performance on the KITTI and EuRoC datasets. This finding highlights the efficacy of line features in enhancing odometry performance in weak-textured scenes. This point is further emphasized by the reconstructed trajectories tested on Seq. floor3\_1, as depicted in Fig. \ref{fig:floor3_1}.

\subsubsection{LiGT Test on Real-world Datasets}
\label{sec:ligt_real}
We further demonstrated the advantages of the proposed LiGT on real-world datasets compared with the original point-based LiGT\cite{cai2021pose}. Rotation is set equal to the ground truth for fair comparisons. Point features and line features were extracted using the method introduced in the outline of Sec. \ref{sec:experiments}. The results are shown in Fig. \ref{fig:LiGT_on_Datasets}. Due to the reduced degree of feature components in the coefficient matrices, as illustrated in Sec. \ref{sec:p_ligt}, the proposed point-based LiGT (P-LiGT) outperforms the original form (Cai-LiGT \cite{cai2021pose}) on almost all sequences. In scenes with poor point feature quality, P-LiGT exhibits greater improvement in translation estimation accuracy compared to Cai-LiGT. For example, Seq. 01 in KITTI dataset has similar textures and features in the sky areas, while Seq. V1\_03 and Seq. V2\_03 in EuRoC dataset have motion blur. In these cases, the observation noises of point features significantly affect the accuracy of Cai-LiGT, but have less effect on the accuracy of P-LiGT. Furthermore, the results of the proposed point-line-based LiGT depicted in Fig. \ref{fig:LiGT_on_Datasets} reaffirm the advantages of point-line fusion. 

\section{Conclusion}
\label{sec:conclusion}                                
In this paper, we propose an accurate and real-time algorithm termed RT$^2$PL for three-view pose estimation. Compared to trifocal-tensor-based methods and PNEC, RT$^2$PL improves the accuracy both in general and degenerate cases. This improvement arises from a decoupled pose estimation with a probability-aware point-line-based rotation estimation and an accurate point-line-based linear translation constraint. Rigorous experiments on synthetic data and real-world data demonstrated the effectiveness of the proposed modules. These experiments also highlighted the potential of RT$^2$PL for more accurate initial pose estimation in visual odometry applications. 
In the future, we plan to apply the proposed RT$^2$PL to visual-inertial odometry systems.

{\appendices
\section{Holt-LiGT Derivation}
\label{app:Holt-LiGT}
We rewrite the relative translation constraint proposed in \cite{holt1994motion} (i.e. Eq. (5) of \cite{holt1994motion}) with a unique symbol representation in this paper as below:
\begin{equation}
    \begin{aligned}
        &({}^1\bold{t}_0^T \bold{R}_{10} {}^0\bold{n})[{}^2\bold{n}^T \bold{R}_{21}([{}^1\bold{n}]_\times^2 \bold{R}_{10}{}^0)] \\
        &=(-{}^2\bold{t}_1^T {}^2\bold{n})[{}^0\bold{n}^T\bold{R}_{10}^([{}^1\bold{n}]_\times^2 \bold{R}_{10}{}^0)].
    \end{aligned}
    \label{eq:holt_constraints}
\end{equation}
Representing the relative translations with global translations, i.e., ${}^1\bold{t}_0 = \bold{R}_{1G}({}^G\bold{t}_0 - {}^G\bold{t}_1)$ and ${}^2\bold{t}_1 = \bold{R}_{2G}({}^G\bold{t}_1 - {}^G\bold{t}_2)$ and substituting them into Eq. (\ref{eq:holt_constraints}), we can get its LiGT form and the coefficient matrices are
\begin{equation}
    \begin{aligned}
        \bold{B} &= -{}^2\bold{n}^T\bold{R}_{21}([{}^1\bold{n}]_\times^2\bold{R}_{10}{}^0\bold{n}){}^0\bold{n}^T\bold{R}_{0G}\\
        \bold{C} &=  
        {}^2\bold{n}^T\bold{R}_{21}([{}^1\bold{n}]_\times^2\bold{R}_{10}{}^0\bold{n}){}^0\bold{n}^T\bold{R}_{0G}\\
        &-{}^0\bold{n}^T\bold{R}_{10}^T([{}^1\bold{n}]_\times^2\bold{R}_{10}{}^0\bold{n}){}^2\bold{n}^T\bold{R}_{2G}\\
        \bold{D} &= -(\bold{B} + \bold{C}).
        \label{eq:Robert_LiGT}
    \end{aligned}
\end{equation}
Therefore, the degrees of the rotation and feature observation components are $3$ and $5$, respectively.
\section{The LiGT with Line-based DPO Constraints}
\begin{figure}[t]
\centering
{
     \includegraphics[width=0.80\linewidth]{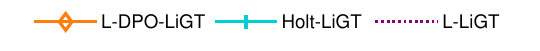}
}
{
\subfigure[rotation error resilience]{
     \includegraphics[width=0.45\linewidth]{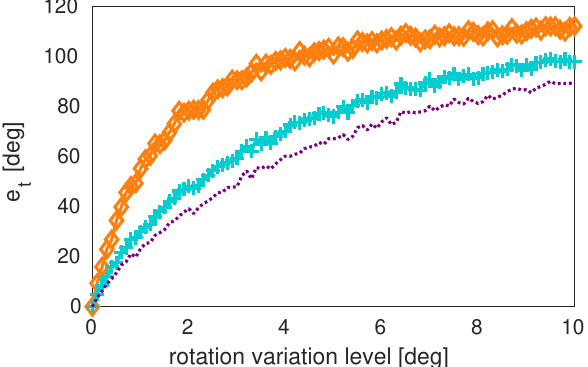}
}
\subfigure[observation noise resilience]{
     \includegraphics[width=0.45\linewidth]{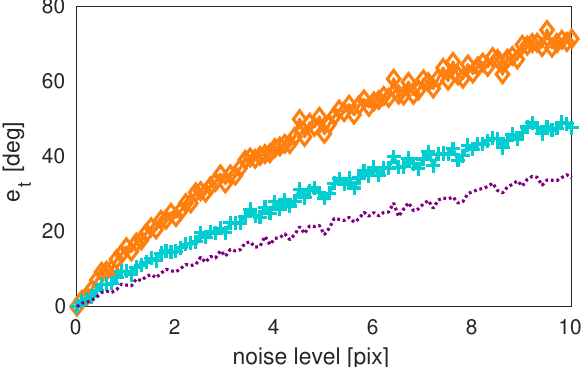}
}
}
   \caption{\textbf{Resilience analysis for Line-based LiGT.} The configuration is the same as that in Sec. \ref{sec:ligt_synthic}.}
   \label{fig:L-DPO-LIGT}
\end{figure}
\label{app:dpo_line}
A 3D spatial line $\mathcal{L}$ in Plücker coordinates is represented by $\mathcal{L} = (z\bold{n}^T, \bold{r}^T)^T$, where $\bold{n}$ is the unit normal vector of the back-projected plane defined as Eq. (\ref{eq:bp_normal}), $\bold{r}$ is the unit direction vector of the line, and $z$ is the distance between the line and the origin of the frame. Given the rotation matrix $\bold{R}_{ji}$ and translation vector ${}^j\bold{t}_i$, we can transform the lücker coordinates of a line from the $i$-th frame to the $j$-th frame \cite{bartoli20013d},
 \begin{equation}
    \begin{bmatrix}
        z_j {}^j\bold{n}\\
        {}^j\bold{r}
    \end{bmatrix} = \begin{bmatrix}
        \bold{R}_{ji} & [{}^j\bold{t}_i]_{\times}\bold{R}_{ji}\\
        \bold{0} & \bold{R}_{ji}
    \end{bmatrix} \begin{bmatrix}
        z_i{}^i\bold{n}\\
        {}^i\bold{r}
    \end{bmatrix}.
     \label{eq: line correspondence}
\end{equation}
According to this geometry transform, we can get the following constraint:
\begin{equation}
    z_j{}^j\bold{n} = z_i\bold{R}_{ji}{}^i\bold{n} - [{}^j\bold{r}]_\times {}^j\bold{t}_i.
    \label{eq: depth correspondence}
\end{equation}
Left multiply the antisymmetric matrix $[{}^j\bold{n}]_\times$ on the both sides of Eq. (\ref{eq: depth correspondence}),
\begin{equation}
    z_i[{}^j\bold{n}]_\times\bold{R}_{ji}{}^i\bold{n} = [{}^j\bold{n}]_\times[{}^j\bold{r}]_\times {}^j\bold{t}_i.
    \label{eq:z_con}
\end{equation}
Taking the magnitude, we get
\begin{equation}
    z_i = \frac{||[{}^j\bold{n}]_\times[{}^j\bold{r}]_\times{}^j\bold{t}_i||}{\theta_{i,j}} \triangleq d^{(i,j)}_i,
    \label{eq: zi}
\end{equation}
where $\theta_{i,j} = ||[{}^j\bold{n}]_\times\bold{R}_{ji}{}^i\bold{n}||$. Similarly, left-multiplying the antisymmetric matrix $[\bold{R}_{ji}{}^i\bold{n}]_\times$ on both sides of Eq. (\ref{eq: depth correspondence}) yields
\begin{equation}
    z_j = \frac{||[\bold{R}_{ji}{}^i\bold{n}]_\times[{}^j\bold{r}]_\times {}^j\bold{t}_i||}{\theta_{i,j}} \triangleq d^{(i,j)}_j.
    \label{eq: zj}
\end{equation}
Combining Eq. (\ref{eq: depth correspondence}), Eq. (\ref{eq: zi}), and Eq. (\ref{eq: zj}), we can obtain 
\begin{equation}
    d^{(i,j)}_j\bold{n}_j=d^{(i,j)}_i\bold{R}_{ji}{}^i\bold{n}-[{}^j\bold{r}]_\times{}^j\bold{t}_i.
\end{equation}
Regarding the $l$-th frame ($l\neq j$), the view pair ($i, l$) also satisfies
\begin{equation}
        d^{(i,l)}_l\bold{n}_l=d^{(i,l)}_i\bold{R}_{li}{}^i\bold{n}-[{}^l\bold{r}]_\times{}^l\bold{t}_i,
        \label{eq:pairi_l}
\end{equation}
and
\begin{equation}
    d^{(i,l)}_i = d^{(i,j)}_i = z_i.
    \label{eq:depth-equal_constraints}
\end{equation}
As \cite{cai2021pose}, we name Eq. (\ref{eq:depth-equal_constraints}) as depth equal constraint of the 3D feature line in $i$-th frame. Substitute Eq. (\ref{eq:depth-equal_constraints}) into Eq. (\ref{eq:pairi_l}),
\begin{equation}
        d^{(i,l)}_l\bold{n}_l=d^{(i,j)}_i\bold{R}_{li}{}^i\bold{n}-[{}^l\bold{r}]_\times{}^l\bold{t}_i.
        \label{eq:pairi_l}
\end{equation}
It is worth noticing that the direction vector $\bold{r}$ is the eigenvector corresponding to the minimal eigenvalue of the matrix $\bold{M}$ (Eq. (\ref{eq:M_contraints})), which is only constructed by line observations and the rotations. Therefore, we name this constraint as DPO constraint for line features. Left-multiplying $[{}^l\bold{n}]_\times$ on both side of the DPO constraint we can get
\begin{equation}
    0 = [{}^l\bold{n}]_\times\left(d^{(i, j)}_i \bold{R}_{li}{}^i\bold{n} 
 - [{}^l\bold{r}]_\times {}^l\bold{t}_i\right)
 \label{eq:ligt_line}
\end{equation}

Left-multiplying $([{}^j\bold{n}]_\times\bold{R}_{ji}{}^i\bold{n})^T$ on Eq. (\ref{eq:z_con}), we get
\begin{equation}
    d^{(i,j)}_i = \frac{\bold{a}_{i,j}^T{}^j\bold{t}_{i}}{\theta_{i,j}^2},
    \label{eq: prop 2}
\end{equation}
where $\bold{a}_{i,j}^T = ([{}^j\bold{n}]_\times\bold{R}_{ji}{}^i\bold{n})^T[{}^j\bold{n}]_\times[{}^j\bold{r}]_\times$. Substituting Eq. (\ref{eq: prop 2}) into Eq. (\ref{eq:ligt_line}), we can get the relative translation constraint:
\begin{equation}
    0 = [{}^l\bold{n}]_\times\left(\bold{R}_{li}{}^i\bold{n}\bold{a}_{i,j}^T{}^j\bold{t}_{i}
 - \theta_{i,j}^2[{}^l\bold{r}]_\times {}^l\bold{t}_i\right).
\end{equation}
Using $0$, $1$, and $2$ replacing $j$, $l$, and $i$, respectively, and representing the relative translations with global translations, we can get its LiGT form and the coefficient matrices are
\begin{equation}
\begin{aligned}
    B &=  - [{}^1\bold{n}]_\times \bold{R}_{12}{}^2\bold{n}\bold{a}^T_{2, 0}\bold{R}_{0G}\\
    C &=  \theta_{2, 0}^2 [{}^1\bold{n}]_\times[{}^1\bold{r}]_\times \bold{R}_{1G}\\
    D &=  - (B+C)
    \label{eq:dpo_line_ligt}
\end{aligned}
\end{equation}
According to Eq. (\ref{eq:dpo_line_ligt}), the degrees of the rotation and feature observation components are over $3$ and $5$, respectively. This is because the direction ${}^1\bold{r}$ is also related to the feature observations and the rotations. We name this LiGT method as \textbf{L-DPO-LiGT} and compare it with L-LiGT and Holt-LiGT on the synthetic data introduced in Sec. \ref{sec:ligt_synthic}. The results are presented in Fig. \ref{fig:L-DPO-LIGT}, which further demonstrate that the lower degree of rotation components and feature components may lead to a more robust LiGT.

}

\bibliographystyle{ieeetr}
\bibliography{IEEEabrv,root}\ 

\begin{thebibliography}{10}

\bibitem{yammine2014novel}
G.~Yammine, E.~Wige, F.~Simmet, D.~Niederkorn, and A.~Kaup, ``Novel similarity-invariant line descriptor and matching algorithm for global motion estimation,'' {\em IEEE Transactions on Circuits and Systems for Video Technology}, vol.~24, no.~8, pp.~1323--1335, 2014.

\bibitem{li2018reliable}
Y.~Li, F.~Wang, R.~Stevenson, R.~Fan, and H.~Tan, ``Reliable line segment matching for multispectral images guided by intersection matches,'' {\em IEEE Transactions on Circuits and Systems for Video Technology}, vol.~29, no.~10, pp.~2899--2912, 2018.

\bibitem{liu1990determination}
Y.~Liu, T.~S. Huang, and O.~D. Faugeras, ``Determination of camera location from 2-d to 3-d line and point correspondences,'' {\em IEEE Transactions on pattern analysis and machine intelligence}, vol.~12, no.~1, pp.~28--37, 1990.

\bibitem{vakhitov2021uncertainty}
A.~Vakhitov, L.~Ferraz, A.~Agudo, and F.~Moreno-Noguer, ``Uncertainty-aware camera pose estimation from points and lines,'' in {\em Proceedings of the IEEE Conference on Computer Vision and Pattern Recognition (CVPR)}, pp.~4659--4668, 2021.

\bibitem{zhang2016comparative}
Y.~Zhang, X.~Li, H.~Liu, and Y.~Shang, ``Comparative study of visual tracking method: A probabilistic approach for pose estimation using lines,'' {\em IEEE Transactions on Circuits and Systems for Video Technology}, vol.~27, no.~6, pp.~1222--1234, 2016.

\bibitem{gomez2019pl}
R.~Gomez-Ojeda, F.-A. Moreno, D.~Zuniga-No{\"e}l, D.~Scaramuzza, and J.~Gonzalez-Jimenez, ``Pl-slam: A stereo slam system through the combination of points and line segments,'' {\em IEEE Transactions on Robotics}, vol.~35, no.~3, pp.~734--746, 2019.

\bibitem{he2018pl}
Y.~He, J.~Zhao, Y.~Guo, W.~He, and K.~Yuan, ``Pl-vio: Tightly-coupled monocular visual--inertial odometry using point and line features,'' {\em Sensors}, vol.~18, no.~4, p.~1159, 2018.

\bibitem{wei2021point}
H.~Wei, F.~Tang, Z.~Xu, C.~Zhang, and Y.~Wu, ``A point-line vio system with novel feature hybrids and with novel line predicting-matching,'' {\em IEEE Robotics and Automation Letters}, vol.~6, no.~4, pp.~8681--8688, 2021.

\bibitem{wei2022structural}
H.~Wei, F.~Tang, Z.~Xu, and Y.~Wu, ``Structural regularity aided visual-inertial odometry with novel coordinate alignment and line triangulation,'' {\em IEEE Robotics and Automation Letters}, vol.~7, no.~4, pp.~10613--10620, 2022.

\bibitem{bartoli2005structure}
A.~Bartoli and P.~Sturm, ``Structure-from-motion using lines: Representation, triangulation, and bundle adjustment,'' {\em Computer Vision and Image Understanding (CVIU)}, vol.~100, no.~3, pp.~416--441, 2005.

\bibitem{hofer2017efficient}
M.~Hofer, M.~Maurer, and H.~Bischof, ``Efficient 3d scene abstraction using line segments,'' {\em Computer Vision and Image Understanding (CVIU)}, vol.~157, pp.~167--178, 2017.

\bibitem{wei2022elsr}
D.~Wei, Y.~Wan, Y.~Zhang, X.~Liu, B.~Zhang, and X.~Wang, ``Elsr: Efficient line segment reconstruction with planes and points guidance,'' in {\em Proceedings of the IEEE Conference on Computer Vision and Pattern Recognition (CVPR)}, pp.~15807--15815, 2022.

\bibitem{liu20233d}
S.~Liu, Y.~Yu, R.~Pautrat, M.~Pollefeys, and V.~Larsson, ``3d line mapping revisited,'' in {\em Proceedings of the IEEE Conference on Computer Vision and Pattern Recognition (CVPR)}, pp.~21445--21455, 2023.

\bibitem{fabbri2020trplp}
R.~Fabbri, T.~Duff, H.~Fan, M.~H. Regan, D.~d. C.~d. Pinho, E.~Tsigaridas, C.~W. Wampler, J.~D. Hauenstein, P.~J. Giblin, B.~Kimia, {\em et~al.}, ``Trplp-trifocal relative pose from lines at points,'' in {\em Proceedings of the IEEE Conference on Computer Vision and Pattern Recognition (CVPR)}, pp.~12073--12083, 2020.

\bibitem{muhle2022probabilistic}
D.~Muhle, L.~Koestler, N.~Demmel, F.~Bernard, and D.~Cremers, ``The probabilistic normal epipolar constraint for frame-to-frame rotation optimization under uncertain feature positions,'' in {\em Proceedings of the IEEE Conference on Computer Vision and Pattern Recognition (CVPR)}, pp.~1819--1828, 2022.

\bibitem{stewenius2006recent}
H.~Stewenius, C.~Engels, and D.~Nist{\'e}r, ``Recent developments on direct relative orientation,'' {\em ISPRS Journal of Photogrammetry and Remote Sensing}, vol.~60, no.~4, pp.~284--294, 2006.

\bibitem{zhang2023cid}
Y.~Zhang, N.~An, C.~Shi, S.~Wang, H.~Wei, P.~Zhang, X.~Meng, Z.~Sun, J.~Wang, W.~Liang, {\em et~al.}, ``Cid-sims: Complex indoor dataset with semantic information and multi-sensor data from a ground wheeled robot viewpoint,'' {\em The International Journal of Robotics Research}, p.~02783649231222507, 2023.

\bibitem{nister2004efficient}
D.~Nist{\'e}r, ``An efficient solution to the five-point relative pose problem,'' {\em IEEE transactions on pattern analysis and machine intelligence}, vol.~26, no.~6, pp.~756--770, 2004.

\bibitem{izquierdo2003estimating}
E.~Izquierdo and V.~Guerra, ``Estimating the essential matrix by efficient linear techniques,'' {\em IEEE Transactions on Circuits and Systems for Video Technology}, vol.~13, no.~9, pp.~925--935, 2003.

\bibitem{kneip2012finding}
L.~Kneip, R.~Siegwart, and M.~Pollefeys, ``Finding the exact rotation between two images independently of the translation,'' in {\em Proceedings of the European Conference on Computer Vision (ECCV)}, pp.~696--709, 2012.

\bibitem{kneip2013direct}
L.~Kneip and S.~Lynen, ``Direct optimization of frame-to-frame rotation,'' in {\em Proceedings of the IEEE International Conference on Computer Vision (ICCV)}, pp.~2352--2359, 2013.

\bibitem{lawson1961contribution}
C.~L. Lawson, ``Contribution to the theory of linear least maximum approximation,'' {\em Ph. D. dissertation. Univ. Calif.}, 1961.

\bibitem{chng2020monocular}
C.-K. Chng, A.~Parra, T.-J. Chin, and Y.~Latif, ``Monocular rotational odometry with incremental rotation averaging and loop closure,'' in {\em 2020 Digital Image Computing: Techniques and Applications (DICTA)}, pp.~1--8, IEEE, 2020.

\bibitem{concha2021instant}
A.~Concha, M.~Burri, J.~Briales, C.~Forster, and L.~Oth, ``Instant visual odometry initialization for mobile ar,'' {\em IEEE Transactions on Visualization and Computer Graphics}, vol.~27, no.~11, pp.~4226--4235, 2021.

\bibitem{he2023rotation}
Y.~He, B.~Xu, Z.~Ouyang, and H.~Li, ``A rotation-translation-decoupled solution for robust and efficient visual-inertial initialization,'' in {\em Proceedings of the IEEE Conference on Computer Vision and Pattern Recognition (CVPR)}, pp.~739--748, 2023.

\bibitem{wang2024stereo}
W.~Wang, C.~Chou, G.~Sevagamoorthy, K.~Chen, Z.~Chen, Z.~Feng, Y.~Xia, F.~Cai, Y.~Xu, and P.~Mordohai, ``Stereo-nec: Enhancing stereo visual-inertial slam initialization with normal epipolar constraints,'' in {\em IEEE International Conference on Robotics and Automation (ICRA)}, 2024.

\bibitem{hartley2003multiple}
R.~Hartley and A.~Zisserman, {\em Multiple view geometry in computer vision}.
\newblock Cambridge university press, 2003.

\bibitem{guan2022trifocal}
B.~Guan, P.~Vasseur, and C.~Demonceaux, ``Trifocal tensor and relative pose estimation from 8 lines and known vertical direction,'' in {\em 2022 IEEE/RSJ International Conference on Intelligent Robots and Systems (IROS)}, pp.~6001--6008, IEEE, 2022.

\bibitem{holt1994motion}
R.~J. Holt and A.~N. Netravali, ``Motion and structure from line correspondences: Some further results,'' {\em International Journal of Imaging Systems and Technology}, vol.~5, no.~1, pp.~52--61, 1994.

\bibitem{hruby2022learning}
P.~Hruby, T.~Duff, A.~Leykin, and T.~Pajdla, ``Learning to solve hard minimal problems,'' in {\em Proceedings of the IEEE Conference on Computer Vision and Pattern Recognition (CVPR)}, pp.~5532--5542, 2022.

\bibitem{duff2019plmp}
T.~Duff, K.~Kohn, A.~Leykin, and T.~Pajdla, ``Plmp-point-line minimal problems in complete multi-view visibility,'' in {\em Proceedings of the IEEE International Conference on Computer Vision (ICCV)}, pp.~1675--1684, 2019.

\bibitem{fabbri2022trifocal}
R.~Fabbri, T.~Duff, H.~Fan, M.~H. Regan, D.~d.~C. de~Pinho, E.~Tsigaridas, C.~W. Wampler, J.~D. Hauenstein, P.~J. Giblin, B.~Kimia, {\em et~al.}, ``Trifocal relative pose from lines at points,'' {\em IEEE Transactions on Pattern Analysis and Machine Intelligence}, 2022.

\bibitem{morgan2009solving}
A.~Morgan, {\em Solving polynomial systems using continuation for engineering and scientific problems}.
\newblock SIAM, 2009.

\bibitem{xu2023plpl}
Z.~Xu, H.~Wei, F.~Tang, Y.~Zhang, Y.~Wu, G.~Ma, S.~Wu, and X.~Jin, ``Plpl-vio: a novel probabilistic line measurement model for point-line-based visual-inertial odometry,'' in {\em 2023 IEEE/RSJ International Conference on Intelligent Robots and Systems (IROS)}, pp.~5211--5218, IEEE, 2023.

\bibitem{uhlmann1995dynamic}
J.~K. Uhlmann, {\em Dynamic map building and localization: new theoretical foundations.}
\newblock PhD thesis, University of Oxford, 1995.

\bibitem{cai2021pose}
Q.~Cai, L.~Zhang, Y.~Wu, W.~Yu, and D.~Hu, ``A pose-only solution to visual reconstruction and navigation,'' {\em IEEE Transactions on Pattern Analysis and Machine Intelligence}, vol.~45, no.~1, pp.~73--86, 2021.

\bibitem{hartley1997defense}
R.~I. Hartley, ``In defense of the eight-point algorithm,'' {\em IEEE Transactions on pattern analysis and machine intelligence}, vol.~19, no.~6, pp.~580--593, 1997.

\bibitem{fischler1981random}
M.~A. Fischler and R.~C. Bolles, ``Random sample consensus: a paradigm for model fitting with applications to image analysis and automated cartography,'' {\em Communications of the ACM}, vol.~24, no.~6, pp.~381--395, 1981.

\bibitem{geiger2012we}
A.~Geiger, P.~Lenz, and R.~Urtasun, ``Are we ready for autonomous driving? the kitti vision benchmark suite,'' in {\em Proceedings of the IEEE Conference on Computer Vision and Pattern Recognition (CVPR)}, pp.~3354--3361, IEEE, 2012.

\bibitem{burri2016euroc}
M.~Burri, J.~Nikolic, P.~Gohl, T.~Schneider, J.~Rehder, S.~Omari, M.~W. Achtelik, and R.~Siegwart, ``The euroc micro aerial vehicle datasets,'' {\em The International Journal of Robotics Research}, vol.~35, no.~10, pp.~1157--1163, 2016.

\bibitem{rosten2006machine}
E.~Rosten and T.~Drummond, ``Machine learning for high-speed corner detection,'' in {\em Computer Vision--ECCV 2006: 9th European Conference on Computer Vision, Graz, Austria, May 7-13, 2006. Proceedings, Part I 9}, pp.~430--443, Springer, 2006.

\bibitem{lucas1981iterative}
B.~D. Lucas and T.~Kanade, ``An iterative image registration technique with an application to stereo vision,'' in {\em IJCAI'81: 7th international joint conference on Artificial intelligence}, vol.~2, pp.~674--679, 1981.

\bibitem{campos2021orb}
C.~Campos, R.~Elvira, J.~J.~G. Rodr{\'\i}guez, J.~M. Montiel, and J.~D. Tard{\'o}s, ``Orb-slam3: An accurate open-source library for visual, visual--inertial, and multimap slam,'' {\em IEEE Transactions on Robotics}, vol.~37, no.~6, pp.~1874--1890, 2021.

\bibitem{von2012lsd}
R.~G. Von~Gioi, J.~Jakubowicz, J.-M. Morel, and G.~Randall, ``Lsd: A line segment detector,'' {\em Image Processing On Line}, vol.~2, pp.~35--55, 2012.

\bibitem{zhang2013efficient}
L.~Zhang and R.~Koch, ``An efficient and robust line segment matching approach based on lbd descriptor and pairwise geometric consistency,'' {\em Journal of visual communication and image representation}, vol.~24, no.~7, pp.~794--805, 2013.

\bibitem{bartoli20013d}
A.~Bartoli and P.~Sturm, ``The 3d line motion matrix and alignment of line reconstructions,'' in {\em Proceedings of the IEEE Conference on Computer Vision and Pattern Recognition (CVPR)}, vol.~1, pp.~I--I, IEEE, 2001.

\end{thebibliography}

\end{document}